%% file: consolidated.tex
\documentclass[12pt]{article}

\usepackage[T1]{fontenc}
\usepackage{graphicx}
\usepackage{epstopdf}
\usepackage{amsmath}
\usepackage{amssymb}
\usepackage{authblk}
\usepackage{cite}
\usepackage{pgf}
\usepackage{pgfplots, pgfplotstable}
\usetikzlibrary{shapes,decorations,calc}
\usepgfplotslibrary{statistics}
\pgfplotsset{compat=1.13}
\pgfmathdeclarefunction{poiss}{1}{%
  \pgfmathparse{(#1^x)*exp(-#1)/(x!)}%
}
\usepackage{gensymb}

\newcommand\blfootnote[1]{%
  \begingroup
  \renewcommand\thefootnote{}\footnote{#1}%
  \addtocounter{footnote}{-1}%
  \endgroup
}

\graphicspath{{images/}}

\begin{document}

\author{Eliott Coyac}
\author{Vincent Gripon}
\author{Charlotte Langlais}
\author{Claude Berrou}
\affil{Electronics Department\\IMT Atlantique\\name.lastname@imt-atlantique.fr}

\title{Robust Associative Memories Naturally Occuring From Recurrent Hebbian Networks Under Noise}

\maketitle

\begin{abstract}

  The brain is a noisy system subject to energy constraints. These facts are rarely taken into account when modelling artificial neural networks. In this paper, we are interested in demonstrating that those factors can actually lead to the appearance of robust associative memories. We first propose a simplified model of noise in the brain, taking into account synaptic noise and interference from neurons external to the network. When coarsely quantized, we show that this noise can be reduced to insertions and erasures. 
We take a neural network with recurrent modifiable connections, and subject it to noisy external inputs. We introduce an energy usage limitation principle in the network as well as consolidated Hebbian learning, resulting in an incremental processing of inputs. We show that the connections naturally formed correspond to state-of-the-art binary sparse associative memories.

\end{abstract}


\section{Introduction}

Sparse associative memories are known to achieve almost optimal performance as far as memory efficiency is concerned~\cite{gripon2015comparative}. These structures have long been considered more biologically plausible than their indexed counterparts for their ability to access and retrieve content from partial inputs. The most efficient models rely on binary connections. They have been inspired by sparse coding and Hebbian learning. In this paper, we are interested in demonstrating that these memories naturally arise when considering recurrent neural networks under energy and noise constraints, assuming some changes on the classical Hebbian learning process.
\blfootnote{This work was funded in part by the European Research Council under the European Union's Seventh Framework Programme (FP7/2007-2013)/ERC grant agreement n$\degree$ 290901. It was also funded in part by the CominLabs project Neural Communications and the Future \& Rupture program.}

We first are interested in introducing a model of noise, as it is present in the brain. Noise in the brain can be due to irrelevant inputs from other neurons, synaptic failure or other factors such as molecular noise. In this work, we focus on the first two of these factors. We show that this noise can be reduced to insertions and erasures when considered external from the network. One way to address noise concerns consists in refining the Hebbian learning rule.

Hebbian learning has been widely used in neural networks since their beginning,
stating that when two neurons fire together, the connection between the neurons
strengthens. There have been some variations, so as to account for information
decay and preventing the synaptic weights from diverging. We present another
approach of Hebbian learning, strengthening already strong enough connections
and decaying weak connections until they disappear, making all weights
eventually 1 or 0 and obtaining a binary network. This approach allows to create a neural network where there is no need to have precise weights for the connections, a criticism that can be made of some recurrent networks as to their biological plausibility, such as Hopfield networks\cite{hopfield1982neural}. This type of Hebbian learning has already been studied to some extent~\cite{fusi2000spike,gerstner2002mathematical} and is called \emph{consolidated Hebbian learning}. We chose to study a recurrent Hebbian neural network with this approach of Hebbian learning.

Finally we add a last constraint to our network: limited energy consumption. This is in line with biological plausibility~\cite{attwell2001energy, lennie2003cost}. By studying the neural network obtained from consolidated Hebbian learning and this constraint, we see that it possesses error-correcting capabilities and in fact apparents itself to either a Willshaw network~\cite{willshaw1969non} or a Neural Clique network~\cite{gripon2011sparse}, depending how energy usage is limited.

 The outline of the paper is as follows. We first study noise in the brain and how it can be generalized in Section II. In Section III we introduce our neural network model, and the constraints we put on it. Finally, in Section IV, we show how it becomes either a Willshaw network or a Neural Clique network depending on the energy efficiency rule chosen. Section V is the conclusion.

\section{Noise model}

\subsection{Synaptic noise}

Neurons are connected to a multitude of other neurons. Each neuron receives inputs from numerous other neurons, sometimes numbering in the thousands. On the other hand, they have a single axon which branches out to reach a multitude of other \emph{target} neurons, so each neuron is itself an input to a number of many other neurons. In addition to this high connectivity, there is not a single point of contact between a neuron and a target neuron, but several. An axon not only branches to reach multiple neurons, it also branches off into several synapses reaching the same target neuron. 

Generally, the connection between two neurons is comprised of 5 to 25 synapses\cite{branco2009probability}. Synapses are not reliable\cite{branco2009probability,allen1994evaluation}, and the probability of them working typically ranges from $0.2$ to $0.8$ \cite{branco2009probability}. 

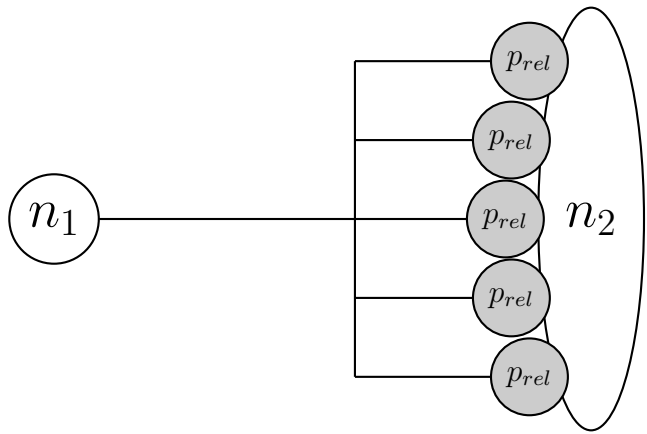
\begin{figure}
  \begin{center}
    \begin{tikzpicture}[thick,darkstyle/.style={circle,draw,fill=gray!40},yscale=0.7]
      
      \node [ellipse, draw,fill=white, font=\LARGE, thick, minimum height=160, minimum width=40] () at (5.14, 0) {$n_2$};
      \draw (-2, 0) -- (2, 0);
      \draw (2, -3) -- (2, 3);
      \foreach \i in {-2,...,2} {
        \draw(2,\i*1.5) -- (4, \i*1.5);
        \node [darkstyle] (\i*1.5) at (4 + \i*\i*0.08, \i*1.5) {$p_{rel}$};
      }
      
      \node [style={circle,draw,fill=white, font=\LARGE}, thick] () at (-2, 0) {$n_1$};
      
    \end{tikzpicture}
  \end{center}
\caption{Example of neuronal contact between one neuron $n_1$ and another $n_2$ with $\mathbf{n_{syn} = 5}$. Each synapse has a probability $\mathbf{p_{rel}}$ of stimulating $n_2$ when $n_1$ is activated. When $n_1$ is activated, $n_2$ receives a stimulation following the law $B(p_{rel}, 5)$.}
\label{fig:synmodel}
\end{figure}

\begin{figure}
  \begin{center}
    \begin{tikzpicture}[
        declare function={binom(\k,\n,\p)=\n!/(\k!*(\n-\k)!)*\p^\k*(1-\p)^(\n-\k);}
      ]
      \begin{axis}[
          samples at={0,...,20},
          yticklabel style={
            /pgf/number format/fixed,
            /pgf/number format/fixed zerofill,
            /pgf/number format/precision=2
          },
          xlabel={Stimulation intensity},
          ylabel={Probability}
        ]
        \addplot [cyan, only marks] {binom(x,20,0.2)}; \addlegendentry{$p_{rel}=0.2$}
        \addplot [orange, mark=x, only marks] {binom(x,20,0.5)}; \addlegendentry{$p_{rel}=0.5$}
      \end{axis}
    \end{tikzpicture}
  \end{center}

  \caption{Probabilities for different stimulation instensities a neuron can receive from another depending on $p_{rel}$ with $n_{syn}=20$.}
\label{fig:binomial}
\end{figure}
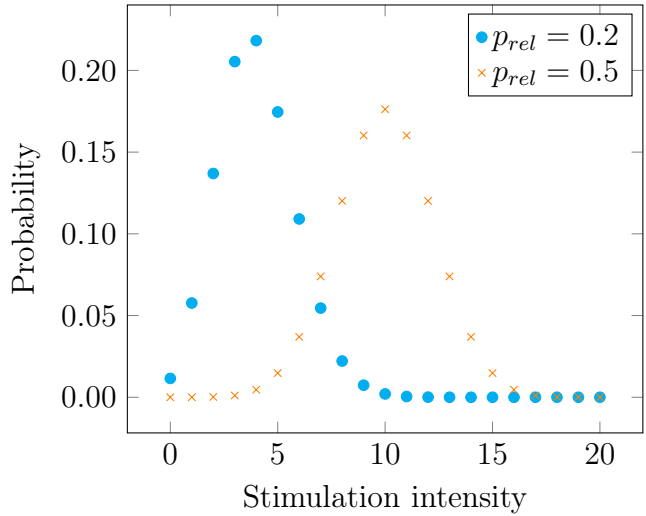

In this paper, we consider the way failing synapses affect the connection between two neurons, and then how this phenomenon contributes to a noisy environment in neural networks. To make things simpler, we consider that any connection between two neurons has the same number of synapses $n_{syn}$, and that the stimulation generated by each synapse is of the same strength. Furthermore we assume that each synapse has the same independent probability $p_{rel}$ of releasing neuro-transmitters when stimulated, regardless of previous events. The connection between two neurons $n_1$ and $n_2$ is represented in Figure~\ref{fig:synmodel}. With this model, the stimulation a neuron receives from another follows a binomial law $B(n_{syn}, p_{rel})$ as shown in Figure~\ref{fig:binomial}.

Synaptic noise interferes with the inner working of a neural network. It is also a modulation on noise generated by external, interfering neurons, which we study in the next subsection. 

\subsection{Interference from other neurons}

Neurons have thousands of inputs and axonal outputs. As such a neural network performing a task can be subject to external input from neurons irrelevant to its current task. We are interested in how the interference caused by these neurons can affect a neural network.

We first focus on quantifying this interference. We assume a neuron $m$ from the network has input from $n_{ext}$ external neurons, not relevant to the neuron's task in the network. We make the hypothesis that those neurons fire at an average frequency of $f_{ext}$ Hz, a typical neuron firing 5 to 50 times per second. 

To have an impact on the neuron $m$, the stimulations from external neurons need to happen in the same short time window $t_{int}$, typically 10 ms. Indeed, we consider the leaky integrator model for the neuron~\cite{jaeger2007optimization}. This is a model where the spikes the neuron receives increase its voltage and the neuron fires once that voltage goes beyond a certain threshold $\sigma$. The neuron's voltage rapidly decreases in the absence of additional stimulations, making it necessary that enough spikes happen in a short time window in order for the neuron to fire.

Finally, we consider that the $n_{ext}$ neurons are composed of $n_{ex}$ excitatory neurons and $n_{in}$ inhibitory neurons, with $n_{in} + n_{ex}$ = $n_{ext}$. We also consider the activations of the $n_{ext}$ external neurons to be completely independent from each other.

Given these parameters we can calculate the average number of excitatory and inhibitory neurons, which $m$ receives stimulation, from during the time period $t_{int}$:

\begin{align}
\lambda_{ex} = n_{ex}\cdot f_{ext} \cdot t_{int}\\
\lambda_{in} = n_{in}\cdot f_{ext} \cdot t_{int}
\end{align}

Those are \emph{averages}. As the interferences are independent with each other, we can use those averages and Poisson's law to calculate the probability that during the time period $t_{int}$ a certain number of inhibitory or excitatory neurons stimulate $m$.

\begin{align}
P_{ex}\left( x \right) = \frac{{e^{ - \lambda_{ex} } \lambda_{ex} ^x }}{{x!}}\\
P_{in}\left( x \right) = \frac{{e^{ - \lambda_{in} } \lambda_{in} ^x }}{{x!}}
\end{align} 

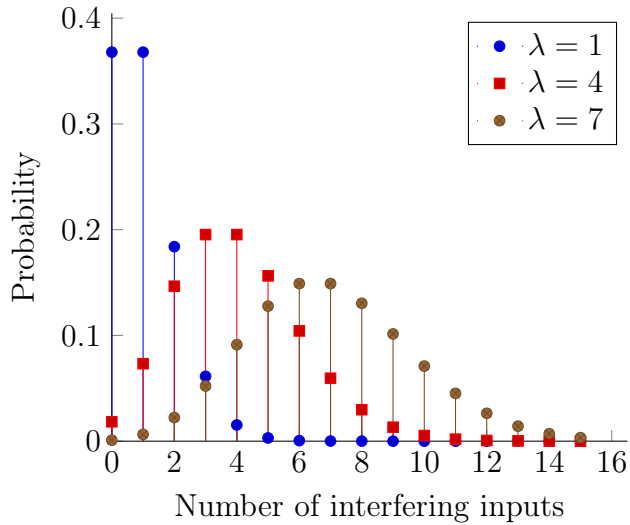
\begin{figure}
  \begin{center}
    \begin{tikzpicture}
      \begin{axis}[every axis plot post/.append style={
            samples at = {0,...,15}},
          axis x line*=bottom,
          axis y line*=left,
          xlabel = {Number of interfering inputs},
          ylabel = {Probability},
          enlargelimits=upper]
        \addplot +[ycomb] {poiss(1)};  \addlegendentry{$\lambda=1$};
        \addplot +[ycomb] {poiss(4)}; \addlegendentry{$\lambda=4$};
        \addplot +[ycomb] {poiss(7)}; \addlegendentry{$\lambda=7$};
      \end{axis}
    \end{tikzpicture}
  \end{center}

  \caption{Different Poisson laws for different average numbers of external neurons interfering. The final interference is a difference of two poisson laws, the one governing the excitatory neurons with the parameter $\lambda_{ex}$, and the one governing the inhibitory neurons with the parameter $\lambda_{in}.$}
\label{fig:poisson}
\end{figure}

Different examples of Poisson laws with parameters are shown in Figure~\ref{fig:poisson}. The figure shows the number of external neurons stimulating $m$. For the exact with the paramaters $p_{rel}$ and $n_{syn}$, a binomial distribution needs to be applied on top of the Poisson distribution.

\subsection{Reduction to insertions and erasures}

Given the threshold $\sigma$ for a neuron to fire, enough positive noise from excitatory neurons can cause a neuron to fire even in the absence of other stimulation. Conversely, enough negative noise from inhibitory neurons can cause a neuron with otherwise enough stimulation to fail to fire.

We call a neuron firing due to positive noise an \emph{insertion}, and a neuron failing to fire due to negative noise an \emph{erasure}. We consider a neural network receiving \emph{external} input. Given the parameters $\sigma$, $p_{rel}$, $n_{syn}$, $\lambda_{ex}$, $\lambda_{in}$, and an additional parameter $n_{inputs}$ which is the number of stimulating neurons feeding a neuron supposed to activate, it is possible to compute $p_{del}$ and $p_{ins}$, the probabilities for erasure and insertion. Each neuron supposed to activate has a probability $p_{del}$ of failing to activate, and each neuron supposed to stay inactive a probability $p_{ins}$ of activating.


More generally, let us consider a model for which in a noisefree environment neurons intended to be receive the same stimulation intensity, and neurons intended to be inactive receive another stimulation intensity. If uniform positive and negative noise factors are introduced to this model, it is possible to reduce those noise factors to two probabilities, a probability of erasure $p_{del}$ and a probability of insertion $p_{ins}$, where each neuron intended to be activated has a probability $p_{del}$ to stay inactive and each neuron intended to stay inactive a probability $p_{ins}$ to be activated.

\section{Network Model}

We consider a neural network of multiple units. Each unit is a mesoscopic coherent group of neurons \cite{galtier2012hebbian}. We choose a network with recurrent modifiable connections, i.e. a plastic network. We also choose our network to be Hebbian \cite{hagan1996neural}, as is the norm.

We consider the network to have $n$ units. The network has several variables that can be observed: the connection weight between units, the external input each unit receives and the activity of each unit. For a time $t$, we note the weight matrix of the network $W(t)$. The connection weight between two units $i$ and $j$  is $W_{ij}(t)$. The activity of a unit $i$ is $V_i(t)$ and $I_i(t)$ is the external input to the unit $i$. 

In the brain, networks typically function at a certain frequency~\cite{buzsaki2004neuronal} . We designate the frequency of our network $f$. As such, we consider the time as discrete, with an increase of $1$ for each iteration of the network.

Given a dataset of $M$ patterns, the network will learn that data by being submitted to each pattern from the dataset for a certain amount of time. More formally, for each pattern in the dataset, the network will be submitted to an external input corresponding to that pattern for a set amount of successive iterations.

\subsection{Consolidated Hebbian learning}

As our network is Hebbian, connections between neurons activating at the same time are strengthened. We choose to strengthen those connections by incrementing them by a set amount $\varepsilon$. Moreover, in order to reduce the impact of the noise, we choose to implement \emph{consolidated Hebbian learning}\cite{fusi2000spike,gerstner2002mathematical}. It consists in Hebbian learning followed by reinforcing strong connections above a threshold $\sigma_{H}$ and weakening weaker connections. It helps ignoring temporary small increase in connections introduced by noise, and maintain strong connections.  

To implement consolidated Hebbian learning, we use a sigmoid $s$ that we apply to each weight of the weight matrix after the Hebbian learning process. 

Any sigmoid in general would fulfill our need, and we use $s$ defined as follows:

\begin{equation}
s(x) \leftarrow \left\{\begin{array}{ll} \dfrac{1}{2} + \dfrac{1}{2}tanh(tan(\pi x-\frac{\pi}{2})) & \mbox {if } x \leq 1, \\
 1 & \mbox {otherwise.}\end{array}\right.
\end{equation}

The reason for this choice of sigmoid is that we want a function with central symmetry with respect to $(0.5,0.5)$. We also want $s(0)=0$, $s(1)=1$ and $s'(0)=s'(1)=0$.

\begin{figure}
  \begin{center}
    \begin{tikzpicture}[
        declare function={sigm(\x)=0.5+0.5*tanh(tan(min(89.9,deg(pi*\x-pi/2))));}
      ]
      \definecolor{darkred}{RGB}{180,10,10}
      
      \begin{axis}[
          xmin=0, xmax=1.1,
          ymin=0, ymax=1.1,
          axis lines=center,
          axis on top=true,
          domain=-0:1,
          ylabel=$y$,
          xlabel=$x$,
        ]

        \addplot [mark=none,draw=red,ultra thick] {sigm(\x)};


        \draw [blue, dotted, thick] (axis cs:+2.5,+1)-- (axis cs:0,+1);
        \draw [blue, dotted, thick] (axis cs:+2.5,+0.5)-- (axis cs:0,+0.5);
        \draw [blue, dotted, thick] (axis cs:+0.5,+0)-- (axis cs:0.5,+1.5);
        \draw [blue, dotted, thick] (axis cs:+1,+0)-- (axis cs:1,+1.5);
        \draw [blue] (axis cs:+0,+0)-- (axis cs:1,+1);
        
        \def\ca{{sigm(0.4)}}
        \pgfmathsetmacro\ca{sigm(0.4)}
        \pgfmathsetmacro\cb{sigm(\ca)}
        \pgfmathsetmacro\cc{sigm(\cb)}
        \pgfmathsetmacro\cd{sigm(\cc)}
        
        \def\da{{sigm(0.55)}}
        \pgfmathsetmacro\da{sigm(0.55)}
        \pgfmathsetmacro\db{sigm(\da)}
        \pgfmathsetmacro\dc{sigm(\db)}
        \pgfmathsetmacro\dd{sigm(\dc)}
        \pgfmathsetmacro\de{sigm(\dd)}
        \pgfmathsetmacro\df{sigm(\de)}
        
        \node[circle, color=darkred] at (0.4,0.4) {\pgfuseplotmark{*}};
        \node at (0.40-0.07,0.40) {A};
        \path[draw=red] (0.4,0.4) -- ( 0.4,\ca) -- (\ca,\ca) -- (\ca,\cb) -- (\cb,\cb) -- (\cb,\cc) 
    	-- (\cc,\cc) -- (\cc,\cd) -- (\cd,\cd) -- (\cd, 0);
        \draw [blue, dashed, thick] (axis cs:+0.4,+0)-- (axis cs:0.4,+\ca);
	
        \node[circle, color=darkred] at (0.55,0.55) {\pgfuseplotmark{*}};
        \node at (0.55-0.07,0.55) {B};
        \path[draw=red] (0.55,0.55) -- (0.55,\da) -- (\da,\da) -- (\da,\db) -- (\db,\db) -- (\db,\dc) 
    	-- (\dc,\dc) -- (\dc,\dd) -- (\dd,\dd) -- (\dd, \de) -- (\de, \de) -- (\de, \df)
    	-- (\df, \df);
        \draw [blue, dashed, thick] (axis cs:+0.55,+0)-- (axis cs:0.55,+0.55);    
        
      \end{axis}
    \end{tikzpicture}
  \end{center}

  \caption{Plotting of the sigmoid $\tfrac{1}{2} + \tfrac{1}{2}tanh(tan(\pi x-\frac{\pi}{2}))$. On the figure we see how an initial connection weight $A$ of $0.4$ is weakened by successive application of the sigmoid until becoming 0, and how an initial connection weight $B$ of $0.55$ is gradually reinforced.}
\label{fig:sigmoid}
\end{figure}
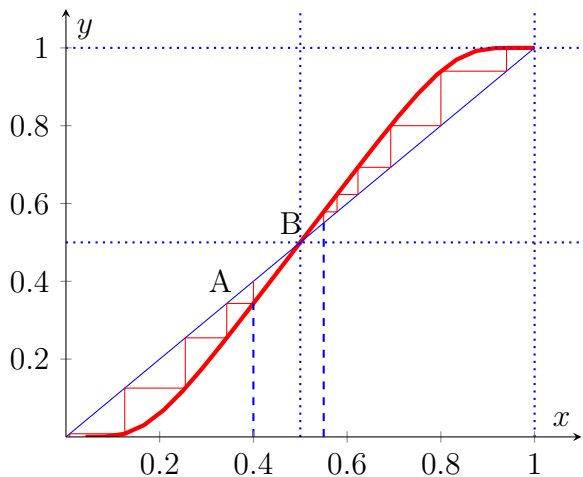

As seen in Figure~\ref{fig:sigmoid}, with this choice for $s$, $\sigma_H = 0.5$. Indeed, after the process of Hebbian learning, each connection weight between two neurons that is above $0.5$ will be reinforced, meaning that eventually they will reach the maximum strength of $1$. Each connection weight below $0.5$ is slightly weakened, and in the absence of other positive stimulations will eventually decrease back to $0$.

\subsection{Limited Energy Usage}

Despite being only $2\%$ of the body mass, the brain makes up for around $20\%$ of the energy consumption of the body~\cite{lennie2003cost}. Energy consumption itself is one of the major limiting factors in the brain~\cite{attwell2001energy, lennie2003cost}. According to~\cite{attwell2001energy}, only one in 100 neurons at most is active at the same time. 

As such, we limit the number of concurrently active units. It can be considered as implementing inhibition in our neural network, as well as assuaging concerns over energy usage. We suggest two ways of implementing this limitation.

\subsubsection{Setting a hard limit without a structure in the network}

The first approach is setting a hard limit to the number of units active at the same time.

We introduce $c$, the limit on the number of units concurrently active in the network. The $c$ units with the highest stimulation, while all the other are forced to be inactive.

Given a group $X$ of $\ell$ units, we introduce a function $Max_c(X)$ which gives the $c$ units with the highest stimulation of $X$.

We also introduce the function $h_c(x)$ for $x \in X$, defined as such:

\begin{equation}
\begin{split}
h_c(x) \leftarrow \left\{\begin{array}{ll} 1 & \mbox {if } x \in Max_c(X), \\
 0 & \mbox {otherwise.}\end{array}\right.
\end{split}
\end{equation}

\subsubsection{Setting a hard limit with a clustered structure in the network}

The second approach is dividing the network into $c$ clusters of $\ell$ units each, and allowing only one neuron per cluster to be active.

For each cluster $X$, a unit $x \in X$ is active if $h_1(x) = 1$.

\medskip


Both this approach and the previous one are a form of \emph{winner-takes-all}, which is widely used in neural networks \cite{wolpert1992stacked,dlugosz2010realization}. Our preference goes to the second approach, as the rule is localized and as such seems more easily enforceable in the brain. It considers that a unit activating inhibits all the other units in the same cluster. Whereas for the first approach, the neural network needs to count the number of units active at the same time, find the $c$ most active units, and inhibits all the other units.



\subsection{Network equations}

With those two rules in place, consolidated Hebbian learning which is there to mitigate noise and the energy usage limitation rule, it is possible to put the network in equations.

Let us denote $H_c$ the winner-takes-all function individually applied to a group of units and $S$ the sigmoid individually applied to each element of a matrix as well.
 
Taking into account the winner-take-all operations on the units of the network and the sigmoid applied to the connection weights, the neural network can be defined by the following set of equations:

\begin{equation}
\left\{\begin{array}{ll} V(t+1) &= H_c\left(W(t) \cdot V(t) + I(t)\right) \\
 W(t+1) &= S\left(\varepsilon \cdot V(t) \otimes V(t) + W(t)\right)\end{array}\right.
\end{equation}

Other neural networks, such as in~\cite{galtier2012hebbian}, use similar-looking differential equations. The major difference between this network and the others is that a sigmoid is applied to the synaptic weights after Hebbian learning. Other networks using a sigmoid apply it to the neurons' activity $V$ instead.

In our network, a pattern is learned after the same units are stimulated for several successive iterations from the external input $I(t)$. The needed number of iterations  depends on $\varepsilon$ and the sigmoid $s$, as well as the noise conditions. If the number of iterations is not enough, after a few iterations on the next pattern, the connection weights between the units constituting the pattern would fall back to $0$. Any fortuitous stimulation of two units at the same time during the operation of the network, for example when recovering a partially erased message, will not create lasting connections.

\subsection{Properties of the network}

Properties that are sought after in a Hebbian neural network have been
summarized and studied \cite{gerstner2002mathematical}. They include:

\subsubsection{Locality} The connection between two neurons only
depends on their activity. In our network, the connection weight between two neurons only depends on the activity of the two neurons and the previous state of the connection weight. As such, \emph{Locality} is respected.
\subsubsection{Boundedness} Connection weights are bounded. In our network, the connection weights are \emph{bounded}, as the sigmoid $S$ is applied to them after each iteration, bounding their weights between 0 and 1.
\subsubsection{Long-term stability} Previously learned information is not
lost. In our network, due to reinforcement, once a connection is above a certain threshold, it will never be lost, introducing \emph{long-term stability}.
\subsubsection{Synaptic depression} The weights of the connections can
decrease. In our network, connection weights below $0.5$ will decrease on their own in the absence of stimulation.
\subsubsection{Incremental learning} Learning new data only requires a
short time. In our network, learning data only requires the corresponding external input to be maintained for some time, forging the new neural connections. It has no relation to previous input, and does not require to go through the whole dataset again. As such, the network supports \emph{incremental learning}.
\subsubsection{Competition} If some weights grow, they do so at the expense of others. This is the only aspect mentioned in \cite{gerstner2002mathematical} that we do not fulfill. There is no mechanism to forget data that has been learned with strong stimulations.

\section{Resulting in a neural clique network}

Depending on the chosen energy efficiency rule, our neural network can either become a Neural Clique network~\cite{gripon2011sparse} or a Willshaw network~\cite{willshaw1969non}. If we divide the network into clusters, we obtain a Neural Clique network, otherwise we obtain a Willshaw network. Neural Clique networks being more efficient than Willshaw networks~\cite{gripon2015comparative}, we will focus on them. 

We first introduce cursory facts about Neural Clique networks and then compare them to our network to see that our network does indeed behave as a Neural Clique network with the same $M$ messages stored.

\subsection{Neural Clique networks}

Neural Clique networks\cite{gripon2011sparse} are associative memories efficiently implemented by neural networks. They use binary connections between units, each connection is 0 or 1. The network is divided in $c$ clusters of $\ell$ units each, as in the energy efficiency rule mentioned in section III-B. 

\begin{figure}[!h]
\centering

\resizebox{1\linewidth}{!}{\input{images/simul3.pgf}}
\caption{Error rate when retrieving half-erased messages from a network of $c=8$ clusters and $\ell=256$ units per clusters. The density of the network is the percentage of possible connections in the network that are actually used to store information.}
\label{fig:error}
\end{figure}
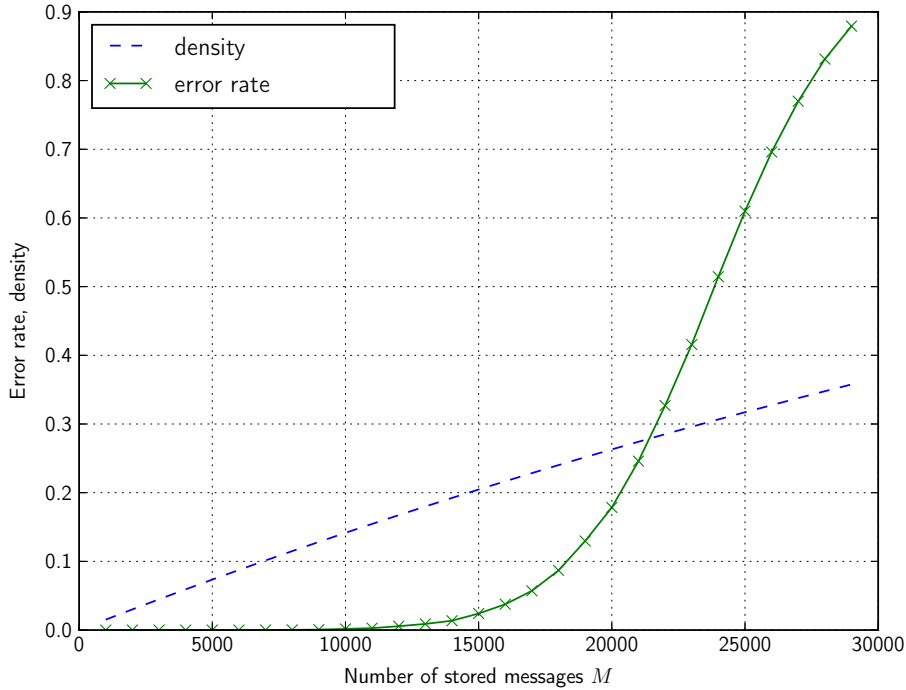

As an exampe of their efficiency, a neural clique network of 2048 units storing messages of 8 bytes can store 15000 messages and retrieve them from half-erased messages with an error rate of less than 2\%, for a storage efficiency of 46\% compared to raw binary storage, as seen in Figure~\ref{fig:error}.

As associative memories, Neural Clique network can test if messages are stored, recover partially erased messages as well as correct errors. 

\subsection{Learning messages}

\begin{figure}[!h]
\centering
\begin{tikzpicture}
  \foreach \type/\fill/\xshift/\yshift in {circle/draw/-50pt/50pt,rectangle/draw/50pt/50pt,rectangle/fill/-50pt/-50pt,circle/fill/50pt/-50pt}{
    \begin{scope}[xshift=\xshift,yshift=\yshift]
    \tikzstyle{every node}=[\type,draw,\fill];
    \foreach \i in {0,20pt,40pt,60pt}{
      \foreach \j in {0,20pt,40pt,60pt}{
        \node (node\type\fill\i\j) at (\i,\j) {};
      }
    }
\end{scope}
  }
  \path
  (nodecircledraw20pt40pt) edge (nodecirclefill040pt)
  edge (noderectanglefill20pt20pt)
  edge (noderectangledraw60pt60pt)
  (nodecirclefill040pt) edge (noderectanglefill20pt20pt)
  edge (noderectangledraw60pt60pt)
  (noderectanglefill20pt20pt) edge (noderectangledraw60pt60pt)
  ;
  \path
  (nodecircledraw60pt60pt) edge (nodecirclefill40pt20pt)
  edge (noderectanglefill060pt)
  edge (noderectangledraw20pt40pt)
  (nodecirclefill40pt20pt) edge (noderectanglefill060pt)
  edge (noderectangledraw20pt40pt)
  (noderectanglefill060pt) edge (noderectangledraw20pt40pt)
  ;
  \path[ultra thick]
  (nodecircledraw020pt) edge (nodecirclefill20pt20pt)
  edge (noderectanglefill40pt60pt)
  edge (noderectangledraw60pt60pt)
  (nodecirclefill20pt20pt) edge (noderectanglefill40pt60pt)
  edge (noderectangledraw60pt60pt)
  (noderectanglefill40pt60pt) edge (noderectangledraw60pt60pt)
  ;
\end{tikzpicture}
\caption{Storing procedure illustration. The pattern to store (with thick edges) connects units from 4 clusters of 16 units each (filled circles, filled rectangles, rectangles and circles).}
\label{fig:motif}
\end{figure}
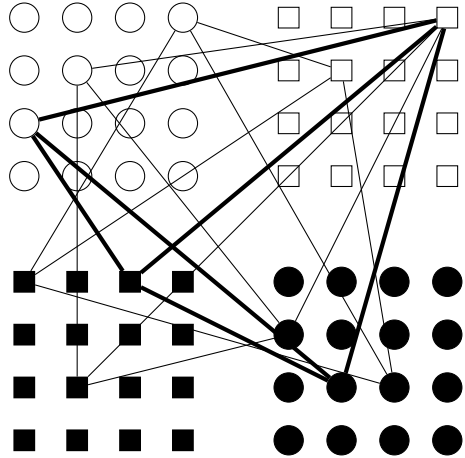

In a Neural Clique network, a message is stored by creating connections between all the corresponding units. An example is shown on Figure~\ref{fig:motif} for a network of $c=4$ clusters of $\ell=16$ units each, for a total of $n=128$. A group of fully interconnected units is also designated as a \emph{clique}, a term from graph theory, hence the name of Neural Clique network.

We consider a Recurrent Hebbian network following our model and using the second energy efficiency rule, and see that in a noisefree environment it reaches the same structure as Neural Clique Network after learning the same patterns. When the network is learning a pattern, the same $c$ units are activated for $n_{iŧ}$ successive iterations. With a proper choice of $\varepsilon$ and $s$, it means that strong enough connections are created between those units. Furthermore no other unit is activated during the learning process, so there is no chance for extra, irregular connections to be formed. As such the network ends up identical to a neural clique network that learned the same patterns.

In this noisefree environment and for the purpose of learning, $\varepsilon$ could be set to $1$, creating instantly a strong connection between two units activated at the same time. This would however cause problems in the decoding process of the network, or if noise is introduced, by creating erroneous connections.

It is also easy to prove that our network can reach arbitrarly close to a Neural Clique Network when the only noise present is erasure noise, and in the last part of this section we provide simulations of what happens when dealing with insertion and erasure noise.

\subsection{Recovering messages}

A common operation in Neural Clique networks is recovering a partially erased message. We describe the decoding process involved.

At first, each unit in the network corresponding to a known part of the message is activated. 

Then each unit \emph{in the whole network} is given a score, which is the number of activated units connected to it. An additional quantity $\gamma$, typically 1, is added if the unit was previously activated. Then a local-winner-takes-all function is applied in each cluster, keeping only the unit with the highest score active. This step is repeated a certain number of times. For example it is typically repeated 5 or 6 times for the parameters $c=8$ and $\ell=256$.

Finally, the result of the operation is the remaining active units. They correspond to the full message. When the message is too degraded, or the network is saturated, the error rate of such an operation increases, as seen in Figure~\ref{fig:error}.

With our network, this process can be achieved by simply activating the units corresponding to the known part of the message, letting the network a few iterations, and picking the still-active units. The only difference is that $\gamma$ is not taken into account. To address that, one way is to create a connection from each unit to itself. Another is to make a neuron more susceptible to be activated if it was already active during the previous iteration.

\subsection{Simulations}

We made the network learn patterns under noisy conditions, and compared its structure to a neural clique network having learned the same messages.

We chose a network of 2048 units, split into $c=8$ clusters of $\ell=256$ units each. The number of successive iterations $n_{it}$ the network is exposed to each message to learn is a parameter. 

We also introduced insertion and erasure noise parameters $p_{ins}=0.05$ and $p_{del}=0.2$. This means that at each iteration, there is on average $(n-c) p_{ins} = 102$ additional units activated in the network, and of the $c=8$ units supposed to be activated on average $ 6.4 $ are actually activated at each iteration. Given the noise parameters, we chose $\varepsilon$ to be $0.18$.

\begin{figure}
  \begin{center}
    \begin{tikzpicture}
      \definecolor{darkgreen}{RGB}{10,160,30}
      
      \begin{axis}[
          xmin=0, xmax=50.1,
          ymin=0, ymax=1.1,
          axis lines=center,
          axis on top=true,
          domain=-0:1,
          ylabel=$w$,
          xlabel=$it$,
        ]
        

        \draw [blue, dotted, thick] (axis cs:+0,+1)-- (axis cs:+50,+1);
        \draw [blue, dotted, thick] (axis cs:+0,+0.5)-- (axis cs:+50,+0.5);

        \path[draw=darkgreen, thick] (0,0) -- (1, 0.0) -- (1, 0.0) -- (2, 0.18) -- (2, 0.04103238310578061) -- (3, 0.2210323831057806) -- (3, 0.0830445122337764) -- (4, 0.2630445122337764) -- (4, 0.1367622351900244) -- (5, 0.1367622351900244) -- (5, 0.012556063893382818) -- (6, 0.012556063893382818) -- (6, 0.0) -- (7, 0.0) -- (7, 0.0) -- (8, 0.18) -- (8, 0.04103238310578061) -- (9, 0.2210323831057806) -- (9, 0.0830445122337764) -- (10, 0.2630445122337764) -- (10, 0.1367622351900244) -- (11, 0.1367622351900244) -- (11, 0.012556063893382818) -- (12, 0.012556063893382818) -- (12, 0.0) -- (13, 0.0) -- (13, 0.0) -- (14, 0.0) -- (14, 0.0) -- (15, 0.18) -- (15, 0.04103238310578061) -- (16, 0.2210323831057806) -- (16, 0.0830445122337764) -- (17, 0.2630445122337764) -- (17, 0.1367622351900244) -- (18, 0.3167622351900244) -- (18, 0.2145082695295783) -- (19, 0.3945082695295783) -- (19, 0.3344321234068682) -- (20, 0.3344321234068682) -- (20, 0.24130949447252387) -- (21, 0.42130949447252386) -- (21, 0.376424648223716) -- (22, 0.376424648223716) -- (22, 0.30619700976903574) -- (23, 0.48619700976903574) -- (23, 0.47831831876047937) -- (24, 0.6583183187604793) -- (24, 0.7475879812751386) -- (25, 0.9275879812751386) -- (25, 0.9998230464147742) -- (26, 0.9998230464147742) -- (26, 1.0) -- (50, 1.0);

        \path[draw=red] (0,0) -- (1, 0.18) -- (1, 0.04103238310578061) -- (2, 0.04103238310578061) -- (2, 1.9918654114992052e-07) -- (3, 1.9918654114992052e-07) -- (3, 0.0) -- (4, 0.18) -- (4, 0.04103238310578061) -- (5, 0.2210323831057806) -- (5, 0.0830445122337764) -- (6, 0.0830445122337764) -- (6, 0.0005576149441937628) -- (7, 0.18055761494419376) -- (7, 0.04151391656555575) -- (8, 0.22151391656555575) -- (8, 0.08360784148629691) -- (9, 0.2636078414862969) -- (9, 0.13753530225675004) -- (10, 0.13753530225675004) -- (10, 0.01290603531351403) -- (11, 0.19290603531351402) -- (11, 0.05285591394167122) -- (12, 0.05285591394167122) -- (12, 6.566436535404563e-06) -- (13, 0.1800065664365354) -- (13, 0.04103803775895981) -- (14, 0.2210380377589598) -- (14, 0.08305111917701036) -- (15, 0.26305111917701035) -- (15, 0.13677129535371296) -- (16, 0.13677129535371296) -- (16, 0.01256013190542582) -- (17, 0.01256013190542582) -- (17, 0.0) -- (18, 0.18) -- (18, 0.04103238310578061) -- (19, 0.2210323831057806) -- (19, 0.0830445122337764) -- (20, 0.0830445122337764) -- (20, 0.0005576149441937628) -- (21, 0.0005576149441937628) -- (21, 0.0) -- (22, 0.18) -- (22, 0.04103238310578061) -- (23, 0.04103238310578061) -- (23, 1.9918654114992052e-07) -- (24, 1.9918654114992052e-07) -- (24, 0.0) -- (25, 0.18) -- (25, 0.04103238310578061) -- (26, 0.2210323831057806) -- (26, 0.0830445122337764) -- (27, 0.0830445122337764) -- (27, 0.0005576149441937628) -- (28, 0.18055761494419376) -- (28, 0.04151391656555575) -- (29, 0.04151391656555575) -- (29, 2.387024674033178e-07) -- (30, 0.1800002387024674) -- (30, 0.041032588656562174) -- (31, 0.22103258865656217) -- (31, 0.08304475239756315) -- (32, 0.08304475239756315) -- (32, 0.0005576275839136446) -- (33, 0.18055762758391364) -- (33, 0.041513927511544424) -- (34, 0.22151392751154442) -- (34, 0.0836078543077392) -- (35, 0.2636078543077392) -- (35, 0.1375353198649058) -- (36, 0.1375353198649058) -- (36, 0.012906043350881857) -- (37, 0.012906043350881857) -- (37, 0.0) -- (38, 0.18) -- (38, 0.04103238310578061) -- (39, 0.2210323831057806) -- (39, 0.0830445122337764) -- (40, 0.0830445122337764) -- (40, 0.0005576149441937628) -- (41, 0.18055761494419376) -- (41, 0.04151391656555575) -- (42, 0.22151391656555575) -- (42, 0.08360784148629691) -- (43, 0.2636078414862969) -- (43, 0.13753530225675004) -- (44, 0.31753530225675003) -- (44, 0.21567186264661714) -- (45, 0.21567186264661714) -- (45, 0.07687116716534259) -- (46, 0.2568711671653426) -- (46, 0.12836801748161236) -- (47, 0.30836801748161236) -- (47, 0.20193550102692825) -- (48, 0.38193550102692825) -- (48, 0.31478945111964624) -- (49, 0.49478945111964623) -- (49, 0.49181528899733684) -- (50, 0.49181528899733684) -- (50, 0.487143486395888);   
        
      \end{axis}
    \end{tikzpicture}
  \end{center}

  \caption{The evolution of connection weight over 50 iterations, under erasure noise. For each iteration, the connection strength is displayed before and after applying the sigmoid. The thick curve depicts a connection that is successfully reinforced beyond $0.5$. The other curve depicts a connection that just fell short being permanent, due to the erasure noise and the limited number of iterations.}
\label{fig:evol}
\end{figure}
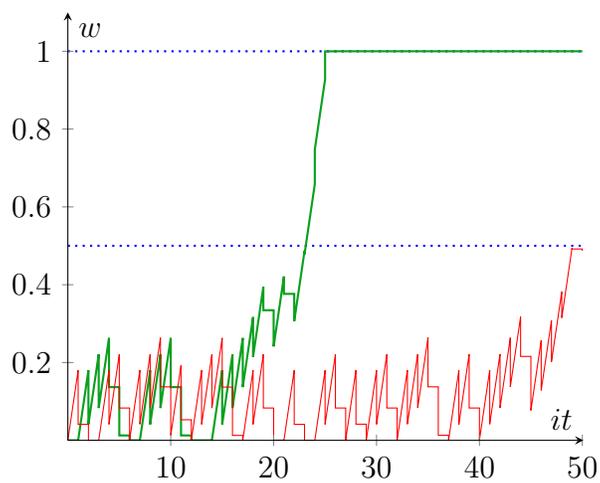

Since the network is Hebbian and a connection can only be strengthened if its two units are activated at the same time, on average only $64\%$ of the relevant connections are actually strengthened on each iteration. Figure~\ref{fig:evol} shows how the weight of the connection between two units can vary over iterations.

A total of $M$ patterns of 8 units are learned by feeding each of them $n_{it}$ successive times to the network in the noise conditions previously described. We then compare the state of the network to a neural clique network in which are stored the same messages. We do the comparison by seeing the difference in connections between the two networks, as the information is stored by the connections. 

The results for different $n_{it}$ are shown in Table~\ref{fig:bac}, where $\mathcal{N}$ is our neural network and the comparisons are in relation to the connections of corresponding Neural Clique network.

\begin{table}[ht!]
\begin{center}
\begin{tabular}{|r|r|r|r|r|}
  \hline
  $n_{it}$ & messages $M$ & connections & added in $\mathcal{N}$ & erased in $\mathcal{N}$ \\
  \hline
  50 & 1000 & 55888 & 0 & 4 \\
  60 & 1000 & 55634 & 4 & 0 \\
  50 & 15000 & 750774 & 12 & 100 \\
  60 & 15000 & 750922 & 52 & 12 \\
  70 & 15000 & 750570 & 72 & 4 \\
  100 & 15000 & 750926 & 86 & 0 \\
  \hline
\end{tabular}
\end{center}
\caption{Connections added and removed in network $\mathcal{N}$ compared to the corresponding neural clique network for different $n_{it}$.}
\label{fig:bac}
\end{table}

We see that the number of connections that differ in both networks is minimal, around $0.005\%$ of the total number of connections of either network. As such our network is virtually the same as a neural clique network in which the same messages are stored. It is possible to adjust $n_{it}$, $\varepsilon$ and $s$ to obtain a number of errors as close to 0 as we want.

\section{Conclusion}

The contributions of this paper are twofold. First we showed that under a few assumptions, all noise in the brain external to a neural network can have its impact on it be reduced to insertions and erasures. Second, by implementing \emph{consolidated Hebbian learning} in order to resist noise and \emph{local-winner-takes-all} in order to limit energy consumption, we showed a recurrent neural network subject to repeated external inputs naturally becomes a neural clique network storing those inputs. 

Due to the properties of a neural clique network, it can be described as a network with a short \emph{incremental} learning period, that can be trivially binarized, and when binarized with an impressive storage capacity that is on the same order of magnitude as binary storage. The recurrent neural network we propose is also governed by a single set of recurent differential equations which are used for both learning new patterns and recovering partially erased messages.

As such, we show that more than simply being bioligically plausible, the presence of neural clique networks in the brain is \emph{probable}, and that a neuro-inspired approach gives birth to a near-optimal associative memory storage-wise.

\bibliographystyle{IEEEtran}
\bibliography{research}

\end{document}

%% file: images/simul3.pgf
\begingroup%
\makeatletter%
\begin{pgfpicture}%
\pgfpathrectangle{\pgfpointorigin}{\pgfqpoint{8.000000in}{6.000000in}}%
\pgfusepath{use as bounding box, clip}%
\begin{pgfscope}%
\pgfsetbuttcap%
\pgfsetmiterjoin%
\definecolor{currentfill}{rgb}{1.000000,1.000000,1.000000}%
\pgfsetfillcolor{currentfill}%
\pgfsetlinewidth{0.000000pt}%
\definecolor{currentstroke}{rgb}{1.000000,1.000000,1.000000}%
\pgfsetstrokecolor{currentstroke}%
\pgfsetdash{}{0pt}%
\pgfpathmoveto{\pgfqpoint{0.000000in}{0.000000in}}%
\pgfpathlineto{\pgfqpoint{8.000000in}{0.000000in}}%
\pgfpathlineto{\pgfqpoint{8.000000in}{6.000000in}}%
\pgfpathlineto{\pgfqpoint{0.000000in}{6.000000in}}%
\pgfpathclose%
\pgfusepath{fill}%
\end{pgfscope}%
\begin{pgfscope}%
\pgfsetbuttcap%
\pgfsetmiterjoin%
\definecolor{currentfill}{rgb}{1.000000,1.000000,1.000000}%
\pgfsetfillcolor{currentfill}%
\pgfsetlinewidth{0.000000pt}%
\definecolor{currentstroke}{rgb}{0.000000,0.000000,0.000000}%
\pgfsetstrokecolor{currentstroke}%
\pgfsetstrokeopacity{0.000000}%
\pgfsetdash{}{0pt}%
\pgfpathmoveto{\pgfqpoint{1.000000in}{0.600000in}}%
\pgfpathlineto{\pgfqpoint{7.200000in}{0.600000in}}%
\pgfpathlineto{\pgfqpoint{7.200000in}{5.400000in}}%
\pgfpathlineto{\pgfqpoint{1.000000in}{5.400000in}}%
\pgfpathclose%
\pgfusepath{fill}%
\end{pgfscope}%
\begin{pgfscope}%
\pgfpathrectangle{\pgfqpoint{1.000000in}{0.600000in}}{\pgfqpoint{6.200000in}{4.800000in}} %
\pgfusepath{clip}%
\pgfsetbuttcap%
\pgfsetroundjoin%
\pgfsetlinewidth{1.003750pt}%
\definecolor{currentstroke}{rgb}{0.000000,0.000000,1.000000}%
\pgfsetstrokecolor{currentstroke}%
\pgfsetdash{{6.000000pt}{6.000000pt}}{0.000000pt}%
\pgfpathmoveto{\pgfqpoint{1.206667in}{0.680784in}}%
\pgfpathlineto{\pgfqpoint{1.413333in}{0.760346in}}%
\pgfpathlineto{\pgfqpoint{1.620000in}{0.838641in}}%
\pgfpathlineto{\pgfqpoint{1.826667in}{0.915797in}}%
\pgfpathlineto{\pgfqpoint{2.033333in}{0.991801in}}%
\pgfpathlineto{\pgfqpoint{2.240000in}{1.066582in}}%
\pgfpathlineto{\pgfqpoint{2.446667in}{1.140317in}}%
\pgfpathlineto{\pgfqpoint{2.653333in}{1.212866in}}%
\pgfpathlineto{\pgfqpoint{2.860000in}{1.284187in}}%
\pgfpathlineto{\pgfqpoint{3.066667in}{1.354967in}}%
\pgfpathlineto{\pgfqpoint{3.273333in}{1.424124in}}%
\pgfpathlineto{\pgfqpoint{3.480000in}{1.492330in}}%
\pgfpathlineto{\pgfqpoint{3.686667in}{1.559503in}}%
\pgfpathlineto{\pgfqpoint{3.893333in}{1.626017in}}%
\pgfpathlineto{\pgfqpoint{4.100000in}{1.691178in}}%
\pgfpathlineto{\pgfqpoint{4.306667in}{1.755197in}}%
\pgfpathlineto{\pgfqpoint{4.513333in}{1.818566in}}%
\pgfpathlineto{\pgfqpoint{4.720000in}{1.881063in}}%
\pgfpathlineto{\pgfqpoint{4.926667in}{1.942171in}}%
\pgfpathlineto{\pgfqpoint{5.133333in}{2.002671in}}%
\pgfpathlineto{\pgfqpoint{5.340000in}{2.062179in}}%
\pgfpathlineto{\pgfqpoint{5.546667in}{2.120741in}}%
\pgfpathlineto{\pgfqpoint{5.753333in}{2.178490in}}%
\pgfpathlineto{\pgfqpoint{5.960000in}{2.235538in}}%
\pgfpathlineto{\pgfqpoint{6.166667in}{2.291347in}}%
\pgfpathlineto{\pgfqpoint{6.373333in}{2.346139in}}%
\pgfpathlineto{\pgfqpoint{6.580000in}{2.401101in}}%
\pgfpathlineto{\pgfqpoint{6.786667in}{2.454660in}}%
\pgfpathlineto{\pgfqpoint{6.993333in}{2.507275in}}%
\pgfusepath{stroke}%
\end{pgfscope}%
\begin{pgfscope}%
\pgfpathrectangle{\pgfqpoint{1.000000in}{0.600000in}}{\pgfqpoint{6.200000in}{4.800000in}} %
\pgfusepath{clip}%
\pgfsetrectcap%
\pgfsetroundjoin%
\pgfsetlinewidth{1.003750pt}%
\definecolor{currentstroke}{rgb}{0.000000,0.500000,0.000000}%
\pgfsetstrokecolor{currentstroke}%
\pgfsetdash{}{0pt}%
\pgfpathmoveto{\pgfqpoint{1.206667in}{0.600000in}}%
\pgfpathlineto{\pgfqpoint{1.413333in}{0.600000in}}%
\pgfpathlineto{\pgfqpoint{1.620000in}{0.600000in}}%
\pgfpathlineto{\pgfqpoint{1.826667in}{0.600000in}}%
\pgfpathlineto{\pgfqpoint{2.033333in}{0.600000in}}%
\pgfpathlineto{\pgfqpoint{2.240000in}{0.600000in}}%
\pgfpathlineto{\pgfqpoint{2.446667in}{0.600800in}}%
\pgfpathlineto{\pgfqpoint{2.653333in}{0.601600in}}%
\pgfpathlineto{\pgfqpoint{2.860000in}{0.603467in}}%
\pgfpathlineto{\pgfqpoint{3.066667in}{0.608800in}}%
\pgfpathlineto{\pgfqpoint{3.273333in}{0.614667in}}%
\pgfpathlineto{\pgfqpoint{3.480000in}{0.629867in}}%
\pgfpathlineto{\pgfqpoint{3.686667in}{0.647733in}}%
\pgfpathlineto{\pgfqpoint{3.893333in}{0.672533in}}%
\pgfpathlineto{\pgfqpoint{4.100000in}{0.728267in}}%
\pgfpathlineto{\pgfqpoint{4.306667in}{0.800267in}}%
\pgfpathlineto{\pgfqpoint{4.513333in}{0.904533in}}%
\pgfpathlineto{\pgfqpoint{4.720000in}{1.063467in}}%
\pgfpathlineto{\pgfqpoint{4.926667in}{1.291467in}}%
\pgfpathlineto{\pgfqpoint{5.133333in}{1.552000in}}%
\pgfpathlineto{\pgfqpoint{5.340000in}{1.911467in}}%
\pgfpathlineto{\pgfqpoint{5.546667in}{2.342667in}}%
\pgfpathlineto{\pgfqpoint{5.753333in}{2.816800in}}%
\pgfpathlineto{\pgfqpoint{5.960000in}{3.344267in}}%
\pgfpathlineto{\pgfqpoint{6.166667in}{3.852800in}}%
\pgfpathlineto{\pgfqpoint{6.373333in}{4.312000in}}%
\pgfpathlineto{\pgfqpoint{6.580000in}{4.706400in}}%
\pgfpathlineto{\pgfqpoint{6.786667in}{5.033867in}}%
\pgfpathlineto{\pgfqpoint{6.993333in}{5.289600in}}%
\pgfusepath{stroke}%
\end{pgfscope}%
\begin{pgfscope}%
\pgfpathrectangle{\pgfqpoint{1.000000in}{0.600000in}}{\pgfqpoint{6.200000in}{4.800000in}} %
\pgfusepath{clip}%
\pgfsetbuttcap%
\pgfsetroundjoin%
\definecolor{currentfill}{rgb}{0.000000,0.500000,0.000000}%
\pgfsetfillcolor{currentfill}%
\pgfsetlinewidth{0.501875pt}%
\definecolor{currentstroke}{rgb}{0.000000,0.500000,0.000000}%
\pgfsetstrokecolor{currentstroke}%
\pgfsetdash{}{0pt}%
\pgfsys@defobject{currentmarker}{\pgfqpoint{-0.041667in}{-0.041667in}}{\pgfqpoint{0.041667in}{0.041667in}}{%
\pgfpathmoveto{\pgfqpoint{-0.041667in}{-0.041667in}}%
\pgfpathlineto{\pgfqpoint{0.041667in}{0.041667in}}%
\pgfpathmoveto{\pgfqpoint{-0.041667in}{0.041667in}}%
\pgfpathlineto{\pgfqpoint{0.041667in}{-0.041667in}}%
\pgfusepath{stroke,fill}%
}%
\begin{pgfscope}%
\pgfsys@transformshift{1.206667in}{0.600000in}%
\pgfsys@useobject{currentmarker}{}%
\end{pgfscope}%
\begin{pgfscope}%
\pgfsys@transformshift{1.413333in}{0.600000in}%
\pgfsys@useobject{currentmarker}{}%
\end{pgfscope}%
\begin{pgfscope}%
\pgfsys@transformshift{1.620000in}{0.600000in}%
\pgfsys@useobject{currentmarker}{}%
\end{pgfscope}%
\begin{pgfscope}%
\pgfsys@transformshift{1.826667in}{0.600000in}%
\pgfsys@useobject{currentmarker}{}%
\end{pgfscope}%
\begin{pgfscope}%
\pgfsys@transformshift{2.033333in}{0.600000in}%
\pgfsys@useobject{currentmarker}{}%
\end{pgfscope}%
\begin{pgfscope}%
\pgfsys@transformshift{2.240000in}{0.600000in}%
\pgfsys@useobject{currentmarker}{}%
\end{pgfscope}%
\begin{pgfscope}%
\pgfsys@transformshift{2.446667in}{0.600800in}%
\pgfsys@useobject{currentmarker}{}%
\end{pgfscope}%
\begin{pgfscope}%
\pgfsys@transformshift{2.653333in}{0.601600in}%
\pgfsys@useobject{currentmarker}{}%
\end{pgfscope}%
\begin{pgfscope}%
\pgfsys@transformshift{2.860000in}{0.603467in}%
\pgfsys@useobject{currentmarker}{}%
\end{pgfscope}%
\begin{pgfscope}%
\pgfsys@transformshift{3.066667in}{0.608800in}%
\pgfsys@useobject{currentmarker}{}%
\end{pgfscope}%
\begin{pgfscope}%
\pgfsys@transformshift{3.273333in}{0.614667in}%
\pgfsys@useobject{currentmarker}{}%
\end{pgfscope}%
\begin{pgfscope}%
\pgfsys@transformshift{3.480000in}{0.629867in}%
\pgfsys@useobject{currentmarker}{}%
\end{pgfscope}%
\begin{pgfscope}%
\pgfsys@transformshift{3.686667in}{0.647733in}%
\pgfsys@useobject{currentmarker}{}%
\end{pgfscope}%
\begin{pgfscope}%
\pgfsys@transformshift{3.893333in}{0.672533in}%
\pgfsys@useobject{currentmarker}{}%
\end{pgfscope}%
\begin{pgfscope}%
\pgfsys@transformshift{4.100000in}{0.728267in}%
\pgfsys@useobject{currentmarker}{}%
\end{pgfscope}%
\begin{pgfscope}%
\pgfsys@transformshift{4.306667in}{0.800267in}%
\pgfsys@useobject{currentmarker}{}%
\end{pgfscope}%
\begin{pgfscope}%
\pgfsys@transformshift{4.513333in}{0.904533in}%
\pgfsys@useobject{currentmarker}{}%
\end{pgfscope}%
\begin{pgfscope}%
\pgfsys@transformshift{4.720000in}{1.063467in}%
\pgfsys@useobject{currentmarker}{}%
\end{pgfscope}%
\begin{pgfscope}%
\pgfsys@transformshift{4.926667in}{1.291467in}%
\pgfsys@useobject{currentmarker}{}%
\end{pgfscope}%
\begin{pgfscope}%
\pgfsys@transformshift{5.133333in}{1.552000in}%
\pgfsys@useobject{currentmarker}{}%
\end{pgfscope}%
\begin{pgfscope}%
\pgfsys@transformshift{5.340000in}{1.911467in}%
\pgfsys@useobject{currentmarker}{}%
\end{pgfscope}%
\begin{pgfscope}%
\pgfsys@transformshift{5.546667in}{2.342667in}%
\pgfsys@useobject{currentmarker}{}%
\end{pgfscope}%
\begin{pgfscope}%
\pgfsys@transformshift{5.753333in}{2.816800in}%
\pgfsys@useobject{currentmarker}{}%
\end{pgfscope}%
\begin{pgfscope}%
\pgfsys@transformshift{5.960000in}{3.344267in}%
\pgfsys@useobject{currentmarker}{}%
\end{pgfscope}%
\begin{pgfscope}%
\pgfsys@transformshift{6.166667in}{3.852800in}%
\pgfsys@useobject{currentmarker}{}%
\end{pgfscope}%
\begin{pgfscope}%
\pgfsys@transformshift{6.373333in}{4.312000in}%
\pgfsys@useobject{currentmarker}{}%
\end{pgfscope}%
\begin{pgfscope}%
\pgfsys@transformshift{6.580000in}{4.706400in}%
\pgfsys@useobject{currentmarker}{}%
\end{pgfscope}%
\begin{pgfscope}%
\pgfsys@transformshift{6.786667in}{5.033867in}%
\pgfsys@useobject{currentmarker}{}%
\end{pgfscope}%
\begin{pgfscope}%
\pgfsys@transformshift{6.993333in}{5.289600in}%
\pgfsys@useobject{currentmarker}{}%
\end{pgfscope}%
\end{pgfscope}%
\begin{pgfscope}%
\pgfsetrectcap%
\pgfsetmiterjoin%
\pgfsetlinewidth{1.003750pt}%
\definecolor{currentstroke}{rgb}{0.000000,0.000000,0.000000}%
\pgfsetstrokecolor{currentstroke}%
\pgfsetdash{}{0pt}%
\pgfpathmoveto{\pgfqpoint{1.000000in}{5.400000in}}%
\pgfpathlineto{\pgfqpoint{7.200000in}{5.400000in}}%
\pgfusepath{stroke}%
\end{pgfscope}%
\begin{pgfscope}%
\pgfsetrectcap%
\pgfsetmiterjoin%
\pgfsetlinewidth{1.003750pt}%
\definecolor{currentstroke}{rgb}{0.000000,0.000000,0.000000}%
\pgfsetstrokecolor{currentstroke}%
\pgfsetdash{}{0pt}%
\pgfpathmoveto{\pgfqpoint{7.200000in}{0.600000in}}%
\pgfpathlineto{\pgfqpoint{7.200000in}{5.400000in}}%
\pgfusepath{stroke}%
\end{pgfscope}%
\begin{pgfscope}%
\pgfsetrectcap%
\pgfsetmiterjoin%
\pgfsetlinewidth{1.003750pt}%
\definecolor{currentstroke}{rgb}{0.000000,0.000000,0.000000}%
\pgfsetstrokecolor{currentstroke}%
\pgfsetdash{}{0pt}%
\pgfpathmoveto{\pgfqpoint{1.000000in}{0.600000in}}%
\pgfpathlineto{\pgfqpoint{7.200000in}{0.600000in}}%
\pgfusepath{stroke}%
\end{pgfscope}%
\begin{pgfscope}%
\pgfsetrectcap%
\pgfsetmiterjoin%
\pgfsetlinewidth{1.003750pt}%
\definecolor{currentstroke}{rgb}{0.000000,0.000000,0.000000}%
\pgfsetstrokecolor{currentstroke}%
\pgfsetdash{}{0pt}%
\pgfpathmoveto{\pgfqpoint{1.000000in}{0.600000in}}%
\pgfpathlineto{\pgfqpoint{1.000000in}{5.400000in}}%
\pgfusepath{stroke}%
\end{pgfscope}%
\begin{pgfscope}%
\pgfpathrectangle{\pgfqpoint{1.000000in}{0.600000in}}{\pgfqpoint{6.200000in}{4.800000in}} %
\pgfusepath{clip}%
\pgfsetbuttcap%
\pgfsetroundjoin%
\pgfsetlinewidth{0.501875pt}%
\definecolor{currentstroke}{rgb}{0.000000,0.000000,0.000000}%
\pgfsetstrokecolor{currentstroke}%
\pgfsetdash{{1.000000pt}{3.000000pt}}{0.000000pt}%
\pgfpathmoveto{\pgfqpoint{1.000000in}{0.600000in}}%
\pgfpathlineto{\pgfqpoint{1.000000in}{5.400000in}}%
\pgfusepath{stroke}%
\end{pgfscope}%
\begin{pgfscope}%
\pgfsetbuttcap%
\pgfsetroundjoin%
\definecolor{currentfill}{rgb}{0.000000,0.000000,0.000000}%
\pgfsetfillcolor{currentfill}%
\pgfsetlinewidth{0.501875pt}%
\definecolor{currentstroke}{rgb}{0.000000,0.000000,0.000000}%
\pgfsetstrokecolor{currentstroke}%
\pgfsetdash{}{0pt}%
\pgfsys@defobject{currentmarker}{\pgfqpoint{0.000000in}{0.000000in}}{\pgfqpoint{0.000000in}{0.055556in}}{%
\pgfpathmoveto{\pgfqpoint{0.000000in}{0.000000in}}%
\pgfpathlineto{\pgfqpoint{0.000000in}{0.055556in}}%
\pgfusepath{stroke,fill}%
}%
\begin{pgfscope}%
\pgfsys@transformshift{1.000000in}{0.600000in}%
\pgfsys@useobject{currentmarker}{}%
\end{pgfscope}%
\end{pgfscope}%
\begin{pgfscope}%
\pgfsetbuttcap%
\pgfsetroundjoin%
\definecolor{currentfill}{rgb}{0.000000,0.000000,0.000000}%
\pgfsetfillcolor{currentfill}%
\pgfsetlinewidth{0.501875pt}%
\definecolor{currentstroke}{rgb}{0.000000,0.000000,0.000000}%
\pgfsetstrokecolor{currentstroke}%
\pgfsetdash{}{0pt}%
\pgfsys@defobject{currentmarker}{\pgfqpoint{0.000000in}{-0.055556in}}{\pgfqpoint{0.000000in}{0.000000in}}{%
\pgfpathmoveto{\pgfqpoint{0.000000in}{0.000000in}}%
\pgfpathlineto{\pgfqpoint{0.000000in}{-0.055556in}}%
\pgfusepath{stroke,fill}%
}%
\begin{pgfscope}%
\pgfsys@transformshift{1.000000in}{5.400000in}%
\pgfsys@useobject{currentmarker}{}%
\end{pgfscope}%
\end{pgfscope}%
\begin{pgfscope}%
\pgftext[x=1.000000in,y=0.544444in,,top]{\sffamily\fontsize{12.000000}{14.400000}\selectfont 0}%
\end{pgfscope}%
\begin{pgfscope}%
\pgfpathrectangle{\pgfqpoint{1.000000in}{0.600000in}}{\pgfqpoint{6.200000in}{4.800000in}} %
\pgfusepath{clip}%
\pgfsetbuttcap%
\pgfsetroundjoin%
\pgfsetlinewidth{0.501875pt}%
\definecolor{currentstroke}{rgb}{0.000000,0.000000,0.000000}%
\pgfsetstrokecolor{currentstroke}%
\pgfsetdash{{1.000000pt}{3.000000pt}}{0.000000pt}%
\pgfpathmoveto{\pgfqpoint{2.033333in}{0.600000in}}%
\pgfpathlineto{\pgfqpoint{2.033333in}{5.400000in}}%
\pgfusepath{stroke}%
\end{pgfscope}%
\begin{pgfscope}%
\pgfsetbuttcap%
\pgfsetroundjoin%
\definecolor{currentfill}{rgb}{0.000000,0.000000,0.000000}%
\pgfsetfillcolor{currentfill}%
\pgfsetlinewidth{0.501875pt}%
\definecolor{currentstroke}{rgb}{0.000000,0.000000,0.000000}%
\pgfsetstrokecolor{currentstroke}%
\pgfsetdash{}{0pt}%
\pgfsys@defobject{currentmarker}{\pgfqpoint{0.000000in}{0.000000in}}{\pgfqpoint{0.000000in}{0.055556in}}{%
\pgfpathmoveto{\pgfqpoint{0.000000in}{0.000000in}}%
\pgfpathlineto{\pgfqpoint{0.000000in}{0.055556in}}%
\pgfusepath{stroke,fill}%
}%
\begin{pgfscope}%
\pgfsys@transformshift{2.033333in}{0.600000in}%
\pgfsys@useobject{currentmarker}{}%
\end{pgfscope}%
\end{pgfscope}%
\begin{pgfscope}%
\pgfsetbuttcap%
\pgfsetroundjoin%
\definecolor{currentfill}{rgb}{0.000000,0.000000,0.000000}%
\pgfsetfillcolor{currentfill}%
\pgfsetlinewidth{0.501875pt}%
\definecolor{currentstroke}{rgb}{0.000000,0.000000,0.000000}%
\pgfsetstrokecolor{currentstroke}%
\pgfsetdash{}{0pt}%
\pgfsys@defobject{currentmarker}{\pgfqpoint{0.000000in}{-0.055556in}}{\pgfqpoint{0.000000in}{0.000000in}}{%
\pgfpathmoveto{\pgfqpoint{0.000000in}{0.000000in}}%
\pgfpathlineto{\pgfqpoint{0.000000in}{-0.055556in}}%
\pgfusepath{stroke,fill}%
}%
\begin{pgfscope}%
\pgfsys@transformshift{2.033333in}{5.400000in}%
\pgfsys@useobject{currentmarker}{}%
\end{pgfscope}%
\end{pgfscope}%
\begin{pgfscope}%
\pgftext[x=2.033333in,y=0.544444in,,top]{\sffamily\fontsize{12.000000}{14.400000}\selectfont 5000}%
\end{pgfscope}%
\begin{pgfscope}%
\pgfpathrectangle{\pgfqpoint{1.000000in}{0.600000in}}{\pgfqpoint{6.200000in}{4.800000in}} %
\pgfusepath{clip}%
\pgfsetbuttcap%
\pgfsetroundjoin%
\pgfsetlinewidth{0.501875pt}%
\definecolor{currentstroke}{rgb}{0.000000,0.000000,0.000000}%
\pgfsetstrokecolor{currentstroke}%
\pgfsetdash{{1.000000pt}{3.000000pt}}{0.000000pt}%
\pgfpathmoveto{\pgfqpoint{3.066667in}{0.600000in}}%
\pgfpathlineto{\pgfqpoint{3.066667in}{5.400000in}}%
\pgfusepath{stroke}%
\end{pgfscope}%
\begin{pgfscope}%
\pgfsetbuttcap%
\pgfsetroundjoin%
\definecolor{currentfill}{rgb}{0.000000,0.000000,0.000000}%
\pgfsetfillcolor{currentfill}%
\pgfsetlinewidth{0.501875pt}%
\definecolor{currentstroke}{rgb}{0.000000,0.000000,0.000000}%
\pgfsetstrokecolor{currentstroke}%
\pgfsetdash{}{0pt}%
\pgfsys@defobject{currentmarker}{\pgfqpoint{0.000000in}{0.000000in}}{\pgfqpoint{0.000000in}{0.055556in}}{%
\pgfpathmoveto{\pgfqpoint{0.000000in}{0.000000in}}%
\pgfpathlineto{\pgfqpoint{0.000000in}{0.055556in}}%
\pgfusepath{stroke,fill}%
}%
\begin{pgfscope}%
\pgfsys@transformshift{3.066667in}{0.600000in}%
\pgfsys@useobject{currentmarker}{}%
\end{pgfscope}%
\end{pgfscope}%
\begin{pgfscope}%
\pgfsetbuttcap%
\pgfsetroundjoin%
\definecolor{currentfill}{rgb}{0.000000,0.000000,0.000000}%
\pgfsetfillcolor{currentfill}%
\pgfsetlinewidth{0.501875pt}%
\definecolor{currentstroke}{rgb}{0.000000,0.000000,0.000000}%
\pgfsetstrokecolor{currentstroke}%
\pgfsetdash{}{0pt}%
\pgfsys@defobject{currentmarker}{\pgfqpoint{0.000000in}{-0.055556in}}{\pgfqpoint{0.000000in}{0.000000in}}{%
\pgfpathmoveto{\pgfqpoint{0.000000in}{0.000000in}}%
\pgfpathlineto{\pgfqpoint{0.000000in}{-0.055556in}}%
\pgfusepath{stroke,fill}%
}%
\begin{pgfscope}%
\pgfsys@transformshift{3.066667in}{5.400000in}%
\pgfsys@useobject{currentmarker}{}%
\end{pgfscope}%
\end{pgfscope}%
\begin{pgfscope}%
\pgftext[x=3.066667in,y=0.544444in,,top]{\sffamily\fontsize{12.000000}{14.400000}\selectfont 10000}%
\end{pgfscope}%
\begin{pgfscope}%
\pgfpathrectangle{\pgfqpoint{1.000000in}{0.600000in}}{\pgfqpoint{6.200000in}{4.800000in}} %
\pgfusepath{clip}%
\pgfsetbuttcap%
\pgfsetroundjoin%
\pgfsetlinewidth{0.501875pt}%
\definecolor{currentstroke}{rgb}{0.000000,0.000000,0.000000}%
\pgfsetstrokecolor{currentstroke}%
\pgfsetdash{{1.000000pt}{3.000000pt}}{0.000000pt}%
\pgfpathmoveto{\pgfqpoint{4.100000in}{0.600000in}}%
\pgfpathlineto{\pgfqpoint{4.100000in}{5.400000in}}%
\pgfusepath{stroke}%
\end{pgfscope}%
\begin{pgfscope}%
\pgfsetbuttcap%
\pgfsetroundjoin%
\definecolor{currentfill}{rgb}{0.000000,0.000000,0.000000}%
\pgfsetfillcolor{currentfill}%
\pgfsetlinewidth{0.501875pt}%
\definecolor{currentstroke}{rgb}{0.000000,0.000000,0.000000}%
\pgfsetstrokecolor{currentstroke}%
\pgfsetdash{}{0pt}%
\pgfsys@defobject{currentmarker}{\pgfqpoint{0.000000in}{0.000000in}}{\pgfqpoint{0.000000in}{0.055556in}}{%
\pgfpathmoveto{\pgfqpoint{0.000000in}{0.000000in}}%
\pgfpathlineto{\pgfqpoint{0.000000in}{0.055556in}}%
\pgfusepath{stroke,fill}%
}%
\begin{pgfscope}%
\pgfsys@transformshift{4.100000in}{0.600000in}%
\pgfsys@useobject{currentmarker}{}%
\end{pgfscope}%
\end{pgfscope}%
\begin{pgfscope}%
\pgfsetbuttcap%
\pgfsetroundjoin%
\definecolor{currentfill}{rgb}{0.000000,0.000000,0.000000}%
\pgfsetfillcolor{currentfill}%
\pgfsetlinewidth{0.501875pt}%
\definecolor{currentstroke}{rgb}{0.000000,0.000000,0.000000}%
\pgfsetstrokecolor{currentstroke}%
\pgfsetdash{}{0pt}%
\pgfsys@defobject{currentmarker}{\pgfqpoint{0.000000in}{-0.055556in}}{\pgfqpoint{0.000000in}{0.000000in}}{%
\pgfpathmoveto{\pgfqpoint{0.000000in}{0.000000in}}%
\pgfpathlineto{\pgfqpoint{0.000000in}{-0.055556in}}%
\pgfusepath{stroke,fill}%
}%
\begin{pgfscope}%
\pgfsys@transformshift{4.100000in}{5.400000in}%
\pgfsys@useobject{currentmarker}{}%
\end{pgfscope}%
\end{pgfscope}%
\begin{pgfscope}%
\pgftext[x=4.100000in,y=0.544444in,,top]{\sffamily\fontsize{12.000000}{14.400000}\selectfont 15000}%
\end{pgfscope}%
\begin{pgfscope}%
\pgfpathrectangle{\pgfqpoint{1.000000in}{0.600000in}}{\pgfqpoint{6.200000in}{4.800000in}} %
\pgfusepath{clip}%
\pgfsetbuttcap%
\pgfsetroundjoin%
\pgfsetlinewidth{0.501875pt}%
\definecolor{currentstroke}{rgb}{0.000000,0.000000,0.000000}%
\pgfsetstrokecolor{currentstroke}%
\pgfsetdash{{1.000000pt}{3.000000pt}}{0.000000pt}%
\pgfpathmoveto{\pgfqpoint{5.133333in}{0.600000in}}%
\pgfpathlineto{\pgfqpoint{5.133333in}{5.400000in}}%
\pgfusepath{stroke}%
\end{pgfscope}%
\begin{pgfscope}%
\pgfsetbuttcap%
\pgfsetroundjoin%
\definecolor{currentfill}{rgb}{0.000000,0.000000,0.000000}%
\pgfsetfillcolor{currentfill}%
\pgfsetlinewidth{0.501875pt}%
\definecolor{currentstroke}{rgb}{0.000000,0.000000,0.000000}%
\pgfsetstrokecolor{currentstroke}%
\pgfsetdash{}{0pt}%
\pgfsys@defobject{currentmarker}{\pgfqpoint{0.000000in}{0.000000in}}{\pgfqpoint{0.000000in}{0.055556in}}{%
\pgfpathmoveto{\pgfqpoint{0.000000in}{0.000000in}}%
\pgfpathlineto{\pgfqpoint{0.000000in}{0.055556in}}%
\pgfusepath{stroke,fill}%
}%
\begin{pgfscope}%
\pgfsys@transformshift{5.133333in}{0.600000in}%
\pgfsys@useobject{currentmarker}{}%
\end{pgfscope}%
\end{pgfscope}%
\begin{pgfscope}%
\pgfsetbuttcap%
\pgfsetroundjoin%
\definecolor{currentfill}{rgb}{0.000000,0.000000,0.000000}%
\pgfsetfillcolor{currentfill}%
\pgfsetlinewidth{0.501875pt}%
\definecolor{currentstroke}{rgb}{0.000000,0.000000,0.000000}%
\pgfsetstrokecolor{currentstroke}%
\pgfsetdash{}{0pt}%
\pgfsys@defobject{currentmarker}{\pgfqpoint{0.000000in}{-0.055556in}}{\pgfqpoint{0.000000in}{0.000000in}}{%
\pgfpathmoveto{\pgfqpoint{0.000000in}{0.000000in}}%
\pgfpathlineto{\pgfqpoint{0.000000in}{-0.055556in}}%
\pgfusepath{stroke,fill}%
}%
\begin{pgfscope}%
\pgfsys@transformshift{5.133333in}{5.400000in}%
\pgfsys@useobject{currentmarker}{}%
\end{pgfscope}%
\end{pgfscope}%
\begin{pgfscope}%
\pgftext[x=5.133333in,y=0.544444in,,top]{\sffamily\fontsize{12.000000}{14.400000}\selectfont 20000}%
\end{pgfscope}%
\begin{pgfscope}%
\pgfpathrectangle{\pgfqpoint{1.000000in}{0.600000in}}{\pgfqpoint{6.200000in}{4.800000in}} %
\pgfusepath{clip}%
\pgfsetbuttcap%
\pgfsetroundjoin%
\pgfsetlinewidth{0.501875pt}%
\definecolor{currentstroke}{rgb}{0.000000,0.000000,0.000000}%
\pgfsetstrokecolor{currentstroke}%
\pgfsetdash{{1.000000pt}{3.000000pt}}{0.000000pt}%
\pgfpathmoveto{\pgfqpoint{6.166667in}{0.600000in}}%
\pgfpathlineto{\pgfqpoint{6.166667in}{5.400000in}}%
\pgfusepath{stroke}%
\end{pgfscope}%
\begin{pgfscope}%
\pgfsetbuttcap%
\pgfsetroundjoin%
\definecolor{currentfill}{rgb}{0.000000,0.000000,0.000000}%
\pgfsetfillcolor{currentfill}%
\pgfsetlinewidth{0.501875pt}%
\definecolor{currentstroke}{rgb}{0.000000,0.000000,0.000000}%
\pgfsetstrokecolor{currentstroke}%
\pgfsetdash{}{0pt}%
\pgfsys@defobject{currentmarker}{\pgfqpoint{0.000000in}{0.000000in}}{\pgfqpoint{0.000000in}{0.055556in}}{%
\pgfpathmoveto{\pgfqpoint{0.000000in}{0.000000in}}%
\pgfpathlineto{\pgfqpoint{0.000000in}{0.055556in}}%
\pgfusepath{stroke,fill}%
}%
\begin{pgfscope}%
\pgfsys@transformshift{6.166667in}{0.600000in}%
\pgfsys@useobject{currentmarker}{}%
\end{pgfscope}%
\end{pgfscope}%
\begin{pgfscope}%
\pgfsetbuttcap%
\pgfsetroundjoin%
\definecolor{currentfill}{rgb}{0.000000,0.000000,0.000000}%
\pgfsetfillcolor{currentfill}%
\pgfsetlinewidth{0.501875pt}%
\definecolor{currentstroke}{rgb}{0.000000,0.000000,0.000000}%
\pgfsetstrokecolor{currentstroke}%
\pgfsetdash{}{0pt}%
\pgfsys@defobject{currentmarker}{\pgfqpoint{0.000000in}{-0.055556in}}{\pgfqpoint{0.000000in}{0.000000in}}{%
\pgfpathmoveto{\pgfqpoint{0.000000in}{0.000000in}}%
\pgfpathlineto{\pgfqpoint{0.000000in}{-0.055556in}}%
\pgfusepath{stroke,fill}%
}%
\begin{pgfscope}%
\pgfsys@transformshift{6.166667in}{5.400000in}%
\pgfsys@useobject{currentmarker}{}%
\end{pgfscope}%
\end{pgfscope}%
\begin{pgfscope}%
\pgftext[x=6.166667in,y=0.544444in,,top]{\sffamily\fontsize{12.000000}{14.400000}\selectfont 25000}%
\end{pgfscope}%
\begin{pgfscope}%
\pgfpathrectangle{\pgfqpoint{1.000000in}{0.600000in}}{\pgfqpoint{6.200000in}{4.800000in}} %
\pgfusepath{clip}%
\pgfsetbuttcap%
\pgfsetroundjoin%
\pgfsetlinewidth{0.501875pt}%
\definecolor{currentstroke}{rgb}{0.000000,0.000000,0.000000}%
\pgfsetstrokecolor{currentstroke}%
\pgfsetdash{{1.000000pt}{3.000000pt}}{0.000000pt}%
\pgfpathmoveto{\pgfqpoint{7.200000in}{0.600000in}}%
\pgfpathlineto{\pgfqpoint{7.200000in}{5.400000in}}%
\pgfusepath{stroke}%
\end{pgfscope}%
\begin{pgfscope}%
\pgfsetbuttcap%
\pgfsetroundjoin%
\definecolor{currentfill}{rgb}{0.000000,0.000000,0.000000}%
\pgfsetfillcolor{currentfill}%
\pgfsetlinewidth{0.501875pt}%
\definecolor{currentstroke}{rgb}{0.000000,0.000000,0.000000}%
\pgfsetstrokecolor{currentstroke}%
\pgfsetdash{}{0pt}%
\pgfsys@defobject{currentmarker}{\pgfqpoint{0.000000in}{0.000000in}}{\pgfqpoint{0.000000in}{0.055556in}}{%
\pgfpathmoveto{\pgfqpoint{0.000000in}{0.000000in}}%
\pgfpathlineto{\pgfqpoint{0.000000in}{0.055556in}}%
\pgfusepath{stroke,fill}%
}%
\begin{pgfscope}%
\pgfsys@transformshift{7.200000in}{0.600000in}%
\pgfsys@useobject{currentmarker}{}%
\end{pgfscope}%
\end{pgfscope}%
\begin{pgfscope}%
\pgfsetbuttcap%
\pgfsetroundjoin%
\definecolor{currentfill}{rgb}{0.000000,0.000000,0.000000}%
\pgfsetfillcolor{currentfill}%
\pgfsetlinewidth{0.501875pt}%
\definecolor{currentstroke}{rgb}{0.000000,0.000000,0.000000}%
\pgfsetstrokecolor{currentstroke}%
\pgfsetdash{}{0pt}%
\pgfsys@defobject{currentmarker}{\pgfqpoint{0.000000in}{-0.055556in}}{\pgfqpoint{0.000000in}{0.000000in}}{%
\pgfpathmoveto{\pgfqpoint{0.000000in}{0.000000in}}%
\pgfpathlineto{\pgfqpoint{0.000000in}{-0.055556in}}%
\pgfusepath{stroke,fill}%
}%
\begin{pgfscope}%
\pgfsys@transformshift{7.200000in}{5.400000in}%
\pgfsys@useobject{currentmarker}{}%
\end{pgfscope}%
\end{pgfscope}%
\begin{pgfscope}%
\pgftext[x=7.200000in,y=0.544444in,,top]{\sffamily\fontsize{12.000000}{14.400000}\selectfont 30000}%
\end{pgfscope}%
\begin{pgfscope}%
\pgftext[x=4.100000in,y=0.313705in,,top]{\sffamily\fontsize{12.000000}{14.400000}\selectfont Number of stored messages $M$}%
\end{pgfscope}%
\begin{pgfscope}%
\pgfpathrectangle{\pgfqpoint{1.000000in}{0.600000in}}{\pgfqpoint{6.200000in}{4.800000in}} %
\pgfusepath{clip}%
\pgfsetbuttcap%
\pgfsetroundjoin%
\pgfsetlinewidth{0.501875pt}%
\definecolor{currentstroke}{rgb}{0.000000,0.000000,0.000000}%
\pgfsetstrokecolor{currentstroke}%
\pgfsetdash{{1.000000pt}{3.000000pt}}{0.000000pt}%
\pgfpathmoveto{\pgfqpoint{1.000000in}{0.600000in}}%
\pgfpathlineto{\pgfqpoint{7.200000in}{0.600000in}}%
\pgfusepath{stroke}%
\end{pgfscope}%
\begin{pgfscope}%
\pgfsetbuttcap%
\pgfsetroundjoin%
\definecolor{currentfill}{rgb}{0.000000,0.000000,0.000000}%
\pgfsetfillcolor{currentfill}%
\pgfsetlinewidth{0.501875pt}%
\definecolor{currentstroke}{rgb}{0.000000,0.000000,0.000000}%
\pgfsetstrokecolor{currentstroke}%
\pgfsetdash{}{0pt}%
\pgfsys@defobject{currentmarker}{\pgfqpoint{0.000000in}{0.000000in}}{\pgfqpoint{0.055556in}{0.000000in}}{%
\pgfpathmoveto{\pgfqpoint{0.000000in}{0.000000in}}%
\pgfpathlineto{\pgfqpoint{0.055556in}{0.000000in}}%
\pgfusepath{stroke,fill}%
}%
\begin{pgfscope}%
\pgfsys@transformshift{1.000000in}{0.600000in}%
\pgfsys@useobject{currentmarker}{}%
\end{pgfscope}%
\end{pgfscope}%
\begin{pgfscope}%
\pgfsetbuttcap%
\pgfsetroundjoin%
\definecolor{currentfill}{rgb}{0.000000,0.000000,0.000000}%
\pgfsetfillcolor{currentfill}%
\pgfsetlinewidth{0.501875pt}%
\definecolor{currentstroke}{rgb}{0.000000,0.000000,0.000000}%
\pgfsetstrokecolor{currentstroke}%
\pgfsetdash{}{0pt}%
\pgfsys@defobject{currentmarker}{\pgfqpoint{-0.055556in}{0.000000in}}{\pgfqpoint{0.000000in}{0.000000in}}{%
\pgfpathmoveto{\pgfqpoint{0.000000in}{0.000000in}}%
\pgfpathlineto{\pgfqpoint{-0.055556in}{0.000000in}}%
\pgfusepath{stroke,fill}%
}%
\begin{pgfscope}%
\pgfsys@transformshift{7.200000in}{0.600000in}%
\pgfsys@useobject{currentmarker}{}%
\end{pgfscope}%
\end{pgfscope}%
\begin{pgfscope}%
\pgftext[x=0.944444in,y=0.600000in,right,]{\sffamily\fontsize{12.000000}{14.400000}\selectfont 0.0}%
\end{pgfscope}%
\begin{pgfscope}%
\pgfpathrectangle{\pgfqpoint{1.000000in}{0.600000in}}{\pgfqpoint{6.200000in}{4.800000in}} %
\pgfusepath{clip}%
\pgfsetbuttcap%
\pgfsetroundjoin%
\pgfsetlinewidth{0.501875pt}%
\definecolor{currentstroke}{rgb}{0.000000,0.000000,0.000000}%
\pgfsetstrokecolor{currentstroke}%
\pgfsetdash{{1.000000pt}{3.000000pt}}{0.000000pt}%
\pgfpathmoveto{\pgfqpoint{1.000000in}{1.133333in}}%
\pgfpathlineto{\pgfqpoint{7.200000in}{1.133333in}}%
\pgfusepath{stroke}%
\end{pgfscope}%
\begin{pgfscope}%
\pgfsetbuttcap%
\pgfsetroundjoin%
\definecolor{currentfill}{rgb}{0.000000,0.000000,0.000000}%
\pgfsetfillcolor{currentfill}%
\pgfsetlinewidth{0.501875pt}%
\definecolor{currentstroke}{rgb}{0.000000,0.000000,0.000000}%
\pgfsetstrokecolor{currentstroke}%
\pgfsetdash{}{0pt}%
\pgfsys@defobject{currentmarker}{\pgfqpoint{0.000000in}{0.000000in}}{\pgfqpoint{0.055556in}{0.000000in}}{%
\pgfpathmoveto{\pgfqpoint{0.000000in}{0.000000in}}%
\pgfpathlineto{\pgfqpoint{0.055556in}{0.000000in}}%
\pgfusepath{stroke,fill}%
}%
\begin{pgfscope}%
\pgfsys@transformshift{1.000000in}{1.133333in}%
\pgfsys@useobject{currentmarker}{}%
\end{pgfscope}%
\end{pgfscope}%
\begin{pgfscope}%
\pgfsetbuttcap%
\pgfsetroundjoin%
\definecolor{currentfill}{rgb}{0.000000,0.000000,0.000000}%
\pgfsetfillcolor{currentfill}%
\pgfsetlinewidth{0.501875pt}%
\definecolor{currentstroke}{rgb}{0.000000,0.000000,0.000000}%
\pgfsetstrokecolor{currentstroke}%
\pgfsetdash{}{0pt}%
\pgfsys@defobject{currentmarker}{\pgfqpoint{-0.055556in}{0.000000in}}{\pgfqpoint{0.000000in}{0.000000in}}{%
\pgfpathmoveto{\pgfqpoint{0.000000in}{0.000000in}}%
\pgfpathlineto{\pgfqpoint{-0.055556in}{0.000000in}}%
\pgfusepath{stroke,fill}%
}%
\begin{pgfscope}%
\pgfsys@transformshift{7.200000in}{1.133333in}%
\pgfsys@useobject{currentmarker}{}%
\end{pgfscope}%
\end{pgfscope}%
\begin{pgfscope}%
\pgftext[x=0.944444in,y=1.133333in,right,]{\sffamily\fontsize{12.000000}{14.400000}\selectfont 0.1}%
\end{pgfscope}%
\begin{pgfscope}%
\pgfpathrectangle{\pgfqpoint{1.000000in}{0.600000in}}{\pgfqpoint{6.200000in}{4.800000in}} %
\pgfusepath{clip}%
\pgfsetbuttcap%
\pgfsetroundjoin%
\pgfsetlinewidth{0.501875pt}%
\definecolor{currentstroke}{rgb}{0.000000,0.000000,0.000000}%
\pgfsetstrokecolor{currentstroke}%
\pgfsetdash{{1.000000pt}{3.000000pt}}{0.000000pt}%
\pgfpathmoveto{\pgfqpoint{1.000000in}{1.666667in}}%
\pgfpathlineto{\pgfqpoint{7.200000in}{1.666667in}}%
\pgfusepath{stroke}%
\end{pgfscope}%
\begin{pgfscope}%
\pgfsetbuttcap%
\pgfsetroundjoin%
\definecolor{currentfill}{rgb}{0.000000,0.000000,0.000000}%
\pgfsetfillcolor{currentfill}%
\pgfsetlinewidth{0.501875pt}%
\definecolor{currentstroke}{rgb}{0.000000,0.000000,0.000000}%
\pgfsetstrokecolor{currentstroke}%
\pgfsetdash{}{0pt}%
\pgfsys@defobject{currentmarker}{\pgfqpoint{0.000000in}{0.000000in}}{\pgfqpoint{0.055556in}{0.000000in}}{%
\pgfpathmoveto{\pgfqpoint{0.000000in}{0.000000in}}%
\pgfpathlineto{\pgfqpoint{0.055556in}{0.000000in}}%
\pgfusepath{stroke,fill}%
}%
\begin{pgfscope}%
\pgfsys@transformshift{1.000000in}{1.666667in}%
\pgfsys@useobject{currentmarker}{}%
\end{pgfscope}%
\end{pgfscope}%
\begin{pgfscope}%
\pgfsetbuttcap%
\pgfsetroundjoin%
\definecolor{currentfill}{rgb}{0.000000,0.000000,0.000000}%
\pgfsetfillcolor{currentfill}%
\pgfsetlinewidth{0.501875pt}%
\definecolor{currentstroke}{rgb}{0.000000,0.000000,0.000000}%
\pgfsetstrokecolor{currentstroke}%
\pgfsetdash{}{0pt}%
\pgfsys@defobject{currentmarker}{\pgfqpoint{-0.055556in}{0.000000in}}{\pgfqpoint{0.000000in}{0.000000in}}{%
\pgfpathmoveto{\pgfqpoint{0.000000in}{0.000000in}}%
\pgfpathlineto{\pgfqpoint{-0.055556in}{0.000000in}}%
\pgfusepath{stroke,fill}%
}%
\begin{pgfscope}%
\pgfsys@transformshift{7.200000in}{1.666667in}%
\pgfsys@useobject{currentmarker}{}%
\end{pgfscope}%
\end{pgfscope}%
\begin{pgfscope}%
\pgftext[x=0.944444in,y=1.666667in,right,]{\sffamily\fontsize{12.000000}{14.400000}\selectfont 0.2}%
\end{pgfscope}%
\begin{pgfscope}%
\pgfpathrectangle{\pgfqpoint{1.000000in}{0.600000in}}{\pgfqpoint{6.200000in}{4.800000in}} %
\pgfusepath{clip}%
\pgfsetbuttcap%
\pgfsetroundjoin%
\pgfsetlinewidth{0.501875pt}%
\definecolor{currentstroke}{rgb}{0.000000,0.000000,0.000000}%
\pgfsetstrokecolor{currentstroke}%
\pgfsetdash{{1.000000pt}{3.000000pt}}{0.000000pt}%
\pgfpathmoveto{\pgfqpoint{1.000000in}{2.200000in}}%
\pgfpathlineto{\pgfqpoint{7.200000in}{2.200000in}}%
\pgfusepath{stroke}%
\end{pgfscope}%
\begin{pgfscope}%
\pgfsetbuttcap%
\pgfsetroundjoin%
\definecolor{currentfill}{rgb}{0.000000,0.000000,0.000000}%
\pgfsetfillcolor{currentfill}%
\pgfsetlinewidth{0.501875pt}%
\definecolor{currentstroke}{rgb}{0.000000,0.000000,0.000000}%
\pgfsetstrokecolor{currentstroke}%
\pgfsetdash{}{0pt}%
\pgfsys@defobject{currentmarker}{\pgfqpoint{0.000000in}{0.000000in}}{\pgfqpoint{0.055556in}{0.000000in}}{%
\pgfpathmoveto{\pgfqpoint{0.000000in}{0.000000in}}%
\pgfpathlineto{\pgfqpoint{0.055556in}{0.000000in}}%
\pgfusepath{stroke,fill}%
}%
\begin{pgfscope}%
\pgfsys@transformshift{1.000000in}{2.200000in}%
\pgfsys@useobject{currentmarker}{}%
\end{pgfscope}%
\end{pgfscope}%
\begin{pgfscope}%
\pgfsetbuttcap%
\pgfsetroundjoin%
\definecolor{currentfill}{rgb}{0.000000,0.000000,0.000000}%
\pgfsetfillcolor{currentfill}%
\pgfsetlinewidth{0.501875pt}%
\definecolor{currentstroke}{rgb}{0.000000,0.000000,0.000000}%
\pgfsetstrokecolor{currentstroke}%
\pgfsetdash{}{0pt}%
\pgfsys@defobject{currentmarker}{\pgfqpoint{-0.055556in}{0.000000in}}{\pgfqpoint{0.000000in}{0.000000in}}{%
\pgfpathmoveto{\pgfqpoint{0.000000in}{0.000000in}}%
\pgfpathlineto{\pgfqpoint{-0.055556in}{0.000000in}}%
\pgfusepath{stroke,fill}%
}%
\begin{pgfscope}%
\pgfsys@transformshift{7.200000in}{2.200000in}%
\pgfsys@useobject{currentmarker}{}%
\end{pgfscope}%
\end{pgfscope}%
\begin{pgfscope}%
\pgftext[x=0.944444in,y=2.200000in,right,]{\sffamily\fontsize{12.000000}{14.400000}\selectfont 0.3}%
\end{pgfscope}%
\begin{pgfscope}%
\pgfpathrectangle{\pgfqpoint{1.000000in}{0.600000in}}{\pgfqpoint{6.200000in}{4.800000in}} %
\pgfusepath{clip}%
\pgfsetbuttcap%
\pgfsetroundjoin%
\pgfsetlinewidth{0.501875pt}%
\definecolor{currentstroke}{rgb}{0.000000,0.000000,0.000000}%
\pgfsetstrokecolor{currentstroke}%
\pgfsetdash{{1.000000pt}{3.000000pt}}{0.000000pt}%
\pgfpathmoveto{\pgfqpoint{1.000000in}{2.733333in}}%
\pgfpathlineto{\pgfqpoint{7.200000in}{2.733333in}}%
\pgfusepath{stroke}%
\end{pgfscope}%
\begin{pgfscope}%
\pgfsetbuttcap%
\pgfsetroundjoin%
\definecolor{currentfill}{rgb}{0.000000,0.000000,0.000000}%
\pgfsetfillcolor{currentfill}%
\pgfsetlinewidth{0.501875pt}%
\definecolor{currentstroke}{rgb}{0.000000,0.000000,0.000000}%
\pgfsetstrokecolor{currentstroke}%
\pgfsetdash{}{0pt}%
\pgfsys@defobject{currentmarker}{\pgfqpoint{0.000000in}{0.000000in}}{\pgfqpoint{0.055556in}{0.000000in}}{%
\pgfpathmoveto{\pgfqpoint{0.000000in}{0.000000in}}%
\pgfpathlineto{\pgfqpoint{0.055556in}{0.000000in}}%
\pgfusepath{stroke,fill}%
}%
\begin{pgfscope}%
\pgfsys@transformshift{1.000000in}{2.733333in}%
\pgfsys@useobject{currentmarker}{}%
\end{pgfscope}%
\end{pgfscope}%
\begin{pgfscope}%
\pgfsetbuttcap%
\pgfsetroundjoin%
\definecolor{currentfill}{rgb}{0.000000,0.000000,0.000000}%
\pgfsetfillcolor{currentfill}%
\pgfsetlinewidth{0.501875pt}%
\definecolor{currentstroke}{rgb}{0.000000,0.000000,0.000000}%
\pgfsetstrokecolor{currentstroke}%
\pgfsetdash{}{0pt}%
\pgfsys@defobject{currentmarker}{\pgfqpoint{-0.055556in}{0.000000in}}{\pgfqpoint{0.000000in}{0.000000in}}{%
\pgfpathmoveto{\pgfqpoint{0.000000in}{0.000000in}}%
\pgfpathlineto{\pgfqpoint{-0.055556in}{0.000000in}}%
\pgfusepath{stroke,fill}%
}%
\begin{pgfscope}%
\pgfsys@transformshift{7.200000in}{2.733333in}%
\pgfsys@useobject{currentmarker}{}%
\end{pgfscope}%
\end{pgfscope}%
\begin{pgfscope}%
\pgftext[x=0.944444in,y=2.733333in,right,]{\sffamily\fontsize{12.000000}{14.400000}\selectfont 0.4}%
\end{pgfscope}%
\begin{pgfscope}%
\pgfpathrectangle{\pgfqpoint{1.000000in}{0.600000in}}{\pgfqpoint{6.200000in}{4.800000in}} %
\pgfusepath{clip}%
\pgfsetbuttcap%
\pgfsetroundjoin%
\pgfsetlinewidth{0.501875pt}%
\definecolor{currentstroke}{rgb}{0.000000,0.000000,0.000000}%
\pgfsetstrokecolor{currentstroke}%
\pgfsetdash{{1.000000pt}{3.000000pt}}{0.000000pt}%
\pgfpathmoveto{\pgfqpoint{1.000000in}{3.266667in}}%
\pgfpathlineto{\pgfqpoint{7.200000in}{3.266667in}}%
\pgfusepath{stroke}%
\end{pgfscope}%
\begin{pgfscope}%
\pgfsetbuttcap%
\pgfsetroundjoin%
\definecolor{currentfill}{rgb}{0.000000,0.000000,0.000000}%
\pgfsetfillcolor{currentfill}%
\pgfsetlinewidth{0.501875pt}%
\definecolor{currentstroke}{rgb}{0.000000,0.000000,0.000000}%
\pgfsetstrokecolor{currentstroke}%
\pgfsetdash{}{0pt}%
\pgfsys@defobject{currentmarker}{\pgfqpoint{0.000000in}{0.000000in}}{\pgfqpoint{0.055556in}{0.000000in}}{%
\pgfpathmoveto{\pgfqpoint{0.000000in}{0.000000in}}%
\pgfpathlineto{\pgfqpoint{0.055556in}{0.000000in}}%
\pgfusepath{stroke,fill}%
}%
\begin{pgfscope}%
\pgfsys@transformshift{1.000000in}{3.266667in}%
\pgfsys@useobject{currentmarker}{}%
\end{pgfscope}%
\end{pgfscope}%
\begin{pgfscope}%
\pgfsetbuttcap%
\pgfsetroundjoin%
\definecolor{currentfill}{rgb}{0.000000,0.000000,0.000000}%
\pgfsetfillcolor{currentfill}%
\pgfsetlinewidth{0.501875pt}%
\definecolor{currentstroke}{rgb}{0.000000,0.000000,0.000000}%
\pgfsetstrokecolor{currentstroke}%
\pgfsetdash{}{0pt}%
\pgfsys@defobject{currentmarker}{\pgfqpoint{-0.055556in}{0.000000in}}{\pgfqpoint{0.000000in}{0.000000in}}{%
\pgfpathmoveto{\pgfqpoint{0.000000in}{0.000000in}}%
\pgfpathlineto{\pgfqpoint{-0.055556in}{0.000000in}}%
\pgfusepath{stroke,fill}%
}%
\begin{pgfscope}%
\pgfsys@transformshift{7.200000in}{3.266667in}%
\pgfsys@useobject{currentmarker}{}%
\end{pgfscope}%
\end{pgfscope}%
\begin{pgfscope}%
\pgftext[x=0.944444in,y=3.266667in,right,]{\sffamily\fontsize{12.000000}{14.400000}\selectfont 0.5}%
\end{pgfscope}%
\begin{pgfscope}%
\pgfpathrectangle{\pgfqpoint{1.000000in}{0.600000in}}{\pgfqpoint{6.200000in}{4.800000in}} %
\pgfusepath{clip}%
\pgfsetbuttcap%
\pgfsetroundjoin%
\pgfsetlinewidth{0.501875pt}%
\definecolor{currentstroke}{rgb}{0.000000,0.000000,0.000000}%
\pgfsetstrokecolor{currentstroke}%
\pgfsetdash{{1.000000pt}{3.000000pt}}{0.000000pt}%
\pgfpathmoveto{\pgfqpoint{1.000000in}{3.800000in}}%
\pgfpathlineto{\pgfqpoint{7.200000in}{3.800000in}}%
\pgfusepath{stroke}%
\end{pgfscope}%
\begin{pgfscope}%
\pgfsetbuttcap%
\pgfsetroundjoin%
\definecolor{currentfill}{rgb}{0.000000,0.000000,0.000000}%
\pgfsetfillcolor{currentfill}%
\pgfsetlinewidth{0.501875pt}%
\definecolor{currentstroke}{rgb}{0.000000,0.000000,0.000000}%
\pgfsetstrokecolor{currentstroke}%
\pgfsetdash{}{0pt}%
\pgfsys@defobject{currentmarker}{\pgfqpoint{0.000000in}{0.000000in}}{\pgfqpoint{0.055556in}{0.000000in}}{%
\pgfpathmoveto{\pgfqpoint{0.000000in}{0.000000in}}%
\pgfpathlineto{\pgfqpoint{0.055556in}{0.000000in}}%
\pgfusepath{stroke,fill}%
}%
\begin{pgfscope}%
\pgfsys@transformshift{1.000000in}{3.800000in}%
\pgfsys@useobject{currentmarker}{}%
\end{pgfscope}%
\end{pgfscope}%
\begin{pgfscope}%
\pgfsetbuttcap%
\pgfsetroundjoin%
\definecolor{currentfill}{rgb}{0.000000,0.000000,0.000000}%
\pgfsetfillcolor{currentfill}%
\pgfsetlinewidth{0.501875pt}%
\definecolor{currentstroke}{rgb}{0.000000,0.000000,0.000000}%
\pgfsetstrokecolor{currentstroke}%
\pgfsetdash{}{0pt}%
\pgfsys@defobject{currentmarker}{\pgfqpoint{-0.055556in}{0.000000in}}{\pgfqpoint{0.000000in}{0.000000in}}{%
\pgfpathmoveto{\pgfqpoint{0.000000in}{0.000000in}}%
\pgfpathlineto{\pgfqpoint{-0.055556in}{0.000000in}}%
\pgfusepath{stroke,fill}%
}%
\begin{pgfscope}%
\pgfsys@transformshift{7.200000in}{3.800000in}%
\pgfsys@useobject{currentmarker}{}%
\end{pgfscope}%
\end{pgfscope}%
\begin{pgfscope}%
\pgftext[x=0.944444in,y=3.800000in,right,]{\sffamily\fontsize{12.000000}{14.400000}\selectfont 0.6}%
\end{pgfscope}%
\begin{pgfscope}%
\pgfpathrectangle{\pgfqpoint{1.000000in}{0.600000in}}{\pgfqpoint{6.200000in}{4.800000in}} %
\pgfusepath{clip}%
\pgfsetbuttcap%
\pgfsetroundjoin%
\pgfsetlinewidth{0.501875pt}%
\definecolor{currentstroke}{rgb}{0.000000,0.000000,0.000000}%
\pgfsetstrokecolor{currentstroke}%
\pgfsetdash{{1.000000pt}{3.000000pt}}{0.000000pt}%
\pgfpathmoveto{\pgfqpoint{1.000000in}{4.333333in}}%
\pgfpathlineto{\pgfqpoint{7.200000in}{4.333333in}}%
\pgfusepath{stroke}%
\end{pgfscope}%
\begin{pgfscope}%
\pgfsetbuttcap%
\pgfsetroundjoin%
\definecolor{currentfill}{rgb}{0.000000,0.000000,0.000000}%
\pgfsetfillcolor{currentfill}%
\pgfsetlinewidth{0.501875pt}%
\definecolor{currentstroke}{rgb}{0.000000,0.000000,0.000000}%
\pgfsetstrokecolor{currentstroke}%
\pgfsetdash{}{0pt}%
\pgfsys@defobject{currentmarker}{\pgfqpoint{0.000000in}{0.000000in}}{\pgfqpoint{0.055556in}{0.000000in}}{%
\pgfpathmoveto{\pgfqpoint{0.000000in}{0.000000in}}%
\pgfpathlineto{\pgfqpoint{0.055556in}{0.000000in}}%
\pgfusepath{stroke,fill}%
}%
\begin{pgfscope}%
\pgfsys@transformshift{1.000000in}{4.333333in}%
\pgfsys@useobject{currentmarker}{}%
\end{pgfscope}%
\end{pgfscope}%
\begin{pgfscope}%
\pgfsetbuttcap%
\pgfsetroundjoin%
\definecolor{currentfill}{rgb}{0.000000,0.000000,0.000000}%
\pgfsetfillcolor{currentfill}%
\pgfsetlinewidth{0.501875pt}%
\definecolor{currentstroke}{rgb}{0.000000,0.000000,0.000000}%
\pgfsetstrokecolor{currentstroke}%
\pgfsetdash{}{0pt}%
\pgfsys@defobject{currentmarker}{\pgfqpoint{-0.055556in}{0.000000in}}{\pgfqpoint{0.000000in}{0.000000in}}{%
\pgfpathmoveto{\pgfqpoint{0.000000in}{0.000000in}}%
\pgfpathlineto{\pgfqpoint{-0.055556in}{0.000000in}}%
\pgfusepath{stroke,fill}%
}%
\begin{pgfscope}%
\pgfsys@transformshift{7.200000in}{4.333333in}%
\pgfsys@useobject{currentmarker}{}%
\end{pgfscope}%
\end{pgfscope}%
\begin{pgfscope}%
\pgftext[x=0.944444in,y=4.333333in,right,]{\sffamily\fontsize{12.000000}{14.400000}\selectfont 0.7}%
\end{pgfscope}%
\begin{pgfscope}%
\pgfpathrectangle{\pgfqpoint{1.000000in}{0.600000in}}{\pgfqpoint{6.200000in}{4.800000in}} %
\pgfusepath{clip}%
\pgfsetbuttcap%
\pgfsetroundjoin%
\pgfsetlinewidth{0.501875pt}%
\definecolor{currentstroke}{rgb}{0.000000,0.000000,0.000000}%
\pgfsetstrokecolor{currentstroke}%
\pgfsetdash{{1.000000pt}{3.000000pt}}{0.000000pt}%
\pgfpathmoveto{\pgfqpoint{1.000000in}{4.866667in}}%
\pgfpathlineto{\pgfqpoint{7.200000in}{4.866667in}}%
\pgfusepath{stroke}%
\end{pgfscope}%
\begin{pgfscope}%
\pgfsetbuttcap%
\pgfsetroundjoin%
\definecolor{currentfill}{rgb}{0.000000,0.000000,0.000000}%
\pgfsetfillcolor{currentfill}%
\pgfsetlinewidth{0.501875pt}%
\definecolor{currentstroke}{rgb}{0.000000,0.000000,0.000000}%
\pgfsetstrokecolor{currentstroke}%
\pgfsetdash{}{0pt}%
\pgfsys@defobject{currentmarker}{\pgfqpoint{0.000000in}{0.000000in}}{\pgfqpoint{0.055556in}{0.000000in}}{%
\pgfpathmoveto{\pgfqpoint{0.000000in}{0.000000in}}%
\pgfpathlineto{\pgfqpoint{0.055556in}{0.000000in}}%
\pgfusepath{stroke,fill}%
}%
\begin{pgfscope}%
\pgfsys@transformshift{1.000000in}{4.866667in}%
\pgfsys@useobject{currentmarker}{}%
\end{pgfscope}%
\end{pgfscope}%
\begin{pgfscope}%
\pgfsetbuttcap%
\pgfsetroundjoin%
\definecolor{currentfill}{rgb}{0.000000,0.000000,0.000000}%
\pgfsetfillcolor{currentfill}%
\pgfsetlinewidth{0.501875pt}%
\definecolor{currentstroke}{rgb}{0.000000,0.000000,0.000000}%
\pgfsetstrokecolor{currentstroke}%
\pgfsetdash{}{0pt}%
\pgfsys@defobject{currentmarker}{\pgfqpoint{-0.055556in}{0.000000in}}{\pgfqpoint{0.000000in}{0.000000in}}{%
\pgfpathmoveto{\pgfqpoint{0.000000in}{0.000000in}}%
\pgfpathlineto{\pgfqpoint{-0.055556in}{0.000000in}}%
\pgfusepath{stroke,fill}%
}%
\begin{pgfscope}%
\pgfsys@transformshift{7.200000in}{4.866667in}%
\pgfsys@useobject{currentmarker}{}%
\end{pgfscope}%
\end{pgfscope}%
\begin{pgfscope}%
\pgftext[x=0.944444in,y=4.866667in,right,]{\sffamily\fontsize{12.000000}{14.400000}\selectfont 0.8}%
\end{pgfscope}%
\begin{pgfscope}%
\pgfpathrectangle{\pgfqpoint{1.000000in}{0.600000in}}{\pgfqpoint{6.200000in}{4.800000in}} %
\pgfusepath{clip}%
\pgfsetbuttcap%
\pgfsetroundjoin%
\pgfsetlinewidth{0.501875pt}%
\definecolor{currentstroke}{rgb}{0.000000,0.000000,0.000000}%
\pgfsetstrokecolor{currentstroke}%
\pgfsetdash{{1.000000pt}{3.000000pt}}{0.000000pt}%
\pgfpathmoveto{\pgfqpoint{1.000000in}{5.400000in}}%
\pgfpathlineto{\pgfqpoint{7.200000in}{5.400000in}}%
\pgfusepath{stroke}%
\end{pgfscope}%
\begin{pgfscope}%
\pgfsetbuttcap%
\pgfsetroundjoin%
\definecolor{currentfill}{rgb}{0.000000,0.000000,0.000000}%
\pgfsetfillcolor{currentfill}%
\pgfsetlinewidth{0.501875pt}%
\definecolor{currentstroke}{rgb}{0.000000,0.000000,0.000000}%
\pgfsetstrokecolor{currentstroke}%
\pgfsetdash{}{0pt}%
\pgfsys@defobject{currentmarker}{\pgfqpoint{0.000000in}{0.000000in}}{\pgfqpoint{0.055556in}{0.000000in}}{%
\pgfpathmoveto{\pgfqpoint{0.000000in}{0.000000in}}%
\pgfpathlineto{\pgfqpoint{0.055556in}{0.000000in}}%
\pgfusepath{stroke,fill}%
}%
\begin{pgfscope}%
\pgfsys@transformshift{1.000000in}{5.400000in}%
\pgfsys@useobject{currentmarker}{}%
\end{pgfscope}%
\end{pgfscope}%
\begin{pgfscope}%
\pgfsetbuttcap%
\pgfsetroundjoin%
\definecolor{currentfill}{rgb}{0.000000,0.000000,0.000000}%
\pgfsetfillcolor{currentfill}%
\pgfsetlinewidth{0.501875pt}%
\definecolor{currentstroke}{rgb}{0.000000,0.000000,0.000000}%
\pgfsetstrokecolor{currentstroke}%
\pgfsetdash{}{0pt}%
\pgfsys@defobject{currentmarker}{\pgfqpoint{-0.055556in}{0.000000in}}{\pgfqpoint{0.000000in}{0.000000in}}{%
\pgfpathmoveto{\pgfqpoint{0.000000in}{0.000000in}}%
\pgfpathlineto{\pgfqpoint{-0.055556in}{0.000000in}}%
\pgfusepath{stroke,fill}%
}%
\begin{pgfscope}%
\pgfsys@transformshift{7.200000in}{5.400000in}%
\pgfsys@useobject{currentmarker}{}%
\end{pgfscope}%
\end{pgfscope}%
\begin{pgfscope}%
\pgftext[x=0.944444in,y=5.400000in,right,]{\sffamily\fontsize{12.000000}{14.400000}\selectfont 0.9}%
\end{pgfscope}%
\begin{pgfscope}%
\pgftext[x=0.609945in,y=3.000000in,,bottom,rotate=90.000000]{\sffamily\fontsize{12.000000}{14.400000}\selectfont Error rate, density}%
\end{pgfscope}%
\begin{pgfscope}%
\pgfsetbuttcap%
\pgfsetmiterjoin%
\definecolor{currentfill}{rgb}{1.000000,1.000000,1.000000}%
\pgfsetfillcolor{currentfill}%
\pgfsetlinewidth{1.003750pt}%
\definecolor{currentstroke}{rgb}{0.000000,0.000000,0.000000}%
\pgfsetstrokecolor{currentstroke}%
\pgfsetdash{}{0pt}%
\pgfpathmoveto{\pgfqpoint{1.100000in}{4.652891in}}%
\pgfpathlineto{\pgfqpoint{3.448320in}{4.652891in}}%
\pgfpathlineto{\pgfqpoint{3.448320in}{5.300000in}}%
\pgfpathlineto{\pgfqpoint{1.100000in}{5.300000in}}%
\pgfpathclose%
\pgfusepath{stroke,fill}%
\end{pgfscope}%
\begin{pgfscope}%
\pgfsetbuttcap%
\pgfsetroundjoin%
\pgfsetlinewidth{1.003750pt}%
\definecolor{currentstroke}{rgb}{0.000000,0.000000,1.000000}%
\pgfsetstrokecolor{currentstroke}%
\pgfsetdash{{6.000000pt}{6.000000pt}}{0.000000pt}%
\pgfpathmoveto{\pgfqpoint{1.240000in}{5.138047in}}%
\pgfpathlineto{\pgfqpoint{1.520000in}{5.138047in}}%
\pgfusepath{stroke}%
\end{pgfscope}%
\begin{pgfscope}%
\pgftext[x=1.740000in,y=5.068047in,left,base]{\sffamily\fontsize{14.400000}{17.280000}\selectfont density}%
\end{pgfscope}%
\begin{pgfscope}%
\pgfsetrectcap%
\pgfsetroundjoin%
\pgfsetlinewidth{1.003750pt}%
\definecolor{currentstroke}{rgb}{0.000000,0.500000,0.000000}%
\pgfsetstrokecolor{currentstroke}%
\pgfsetdash{}{0pt}%
\pgfpathmoveto{\pgfqpoint{1.240000in}{4.844492in}}%
\pgfpathlineto{\pgfqpoint{1.520000in}{4.844492in}}%
\pgfusepath{stroke}%
\end{pgfscope}%
\begin{pgfscope}%
\pgfsetbuttcap%
\pgfsetroundjoin%
\definecolor{currentfill}{rgb}{0.000000,0.500000,0.000000}%
\pgfsetfillcolor{currentfill}%
\pgfsetlinewidth{0.501875pt}%
\definecolor{currentstroke}{rgb}{0.000000,0.500000,0.000000}%
\pgfsetstrokecolor{currentstroke}%
\pgfsetdash{}{0pt}%
\pgfsys@defobject{currentmarker}{\pgfqpoint{-0.041667in}{-0.041667in}}{\pgfqpoint{0.041667in}{0.041667in}}{%
\pgfpathmoveto{\pgfqpoint{-0.041667in}{-0.041667in}}%
\pgfpathlineto{\pgfqpoint{0.041667in}{0.041667in}}%
\pgfpathmoveto{\pgfqpoint{-0.041667in}{0.041667in}}%
\pgfpathlineto{\pgfqpoint{0.041667in}{-0.041667in}}%
\pgfusepath{stroke,fill}%
}%
\begin{pgfscope}%
\pgfsys@transformshift{1.240000in}{4.844492in}%
\pgfsys@useobject{currentmarker}{}%
\end{pgfscope}%
\begin{pgfscope}%
\pgfsys@transformshift{1.520000in}{4.844492in}%
\pgfsys@useobject{currentmarker}{}%
\end{pgfscope}%
\end{pgfscope}%
\begin{pgfscope}%
\pgftext[x=1.740000in,y=4.774492in,left,base]{\sffamily\fontsize{14.400000}{17.280000}\selectfont error rate}%
\end{pgfscope}%
\end{pgfpicture}%
\makeatother%
\endgroup%

%% file: consolidated.bbl
\begin{thebibliography}{10}
\providecommand{\url}[1]{#1}
\csname url@samestyle\endcsname
\providecommand{\newblock}{\relax}
\providecommand{\bibinfo}[2]{#2}
\providecommand{\BIBentrySTDinterwordspacing}{\spaceskip=0pt\relax}
\providecommand{\BIBentryALTinterwordstretchfactor}{4}
\providecommand{\BIBentryALTinterwordspacing}{\spaceskip=\fontdimen2\font plus
\BIBentryALTinterwordstretchfactor\fontdimen3\font minus
  \fontdimen4\font\relax}
\providecommand{\BIBforeignlanguage}[2]{{%
\expandafter\ifx\csname l@#1\endcsname\relax
\typeout{** WARNING: IEEEtran.bst: No hyphenation pattern has been}%
\typeout{** loaded for the language `#1'. Using the pattern for}%
\typeout{** the default language instead.}%
\else
\language=\csname l@#1\endcsname
\fi
#2}}
\providecommand{\BIBdecl}{\relax}
\BIBdecl

\bibitem{gripon2015comparative}
V.~Gripon, J.~Heusel, M.~L{\"o}we, and F.~Vermet, ``A comparative study of
  sparse associative memories,'' \emph{Journal of Statistical Physics}, pp.
  1--25, 2015.

\bibitem{hopfield1982neural}
J.~J. Hopfield, ``Neural networks and physical systems with emergent collective
  computational abilities,'' \emph{Proceedings of the national academy of
  sciences}, vol.~79, no.~8, pp. 2554--2558, 1982.

\bibitem{fusi2000spike}
S.~Fusi, M.~Annunziato, D.~Badoni, A.~Salamon, and D.~J. Amit, ``Spike-driven
  synaptic plasticity: theory, simulation, vlsi implementation,'' \emph{Neural
  Computation}, vol.~12, no.~10, pp. 2227--2258, 2000.

\bibitem{gerstner2002mathematical}
W.~Gerstner and W.~M. Kistler, ``Mathematical formulations of hebbian
  learning,'' \emph{Biological cybernetics}, vol.~87, no. 5-6, pp. 404--415,
  2002.

\bibitem{attwell2001energy}
D.~Attwell and S.~B. Laughlin, ``An energy budget for signaling in the grey
  matter of the brain,'' \emph{Journal of Cerebral Blood Flow \& Metabolism},
  vol.~21, no.~10, pp. 1133--1145, 2001.

\bibitem{lennie2003cost}
P.~Lennie, ``The cost of cortical computation,'' \emph{Current biology},
  vol.~13, no.~6, pp. 493--497, 2003.

\bibitem{willshaw1969non}
D.~J. Willshaw, O.~P. Buneman, and H.~C. Longuet-Higgins, ``Non-holographic
  associative memory.'' \emph{Nature}, 1969.

\bibitem{gripon2011sparse}
V.~Gripon and C.~Berrou, ``Sparse neural networks with large learning
  diversity,'' \emph{Neural Networks, IEEE Transactions on}, vol.~22, no.~7,
  pp. 1087--1096, 2011.

\bibitem{branco2009probability}
T.~Branco and K.~Staras, ``The probability of neurotransmitter release:
  variability and feedback control at single synapses,'' \emph{Nature Reviews
  Neuroscience}, vol.~10, no.~5, pp. 373--383, 2009.

\bibitem{allen1994evaluation}
C.~Allen and C.~F. Stevens, ``An evaluation of causes for unreliability of
  synaptic transmission,'' \emph{Proceedings of the National Academy of
  Sciences}, vol.~91, no.~22, pp. 10\,380--10\,383, 1994.

\bibitem{jaeger2007optimization}
H.~Jaeger, M.~Luko{\v{s}}evi{\v{c}}ius, D.~Popovici, and U.~Siewert,
  ``Optimization and applications of echo state networks with leaky-integrator
  neurons,'' \emph{Neural networks}, vol.~20, no.~3, pp. 335--352, 2007.

\bibitem{galtier2012hebbian}
M.~N. Galtier, O.~D. Faugeras, and P.~C. Bressloff, ``Hebbian learning of
  recurrent connections: a geometrical perspective,'' \emph{Neural
  computation}, vol.~24, no.~9, pp. 2346--2383, 2012.

\bibitem{hagan1996neural}
M.~T. Hagan, H.~B. Demuth, M.~H. Beale, and O.~De~Jes{\'u}s, \emph{Neural
  network design}.\hskip 1em plus 0.5em minus 0.4em\relax PWS publishing
  company Boston, 1996, vol.~20.

\bibitem{buzsaki2004neuronal}
G.~Buzs{\'a}ki and A.~Draguhn, ``Neuronal oscillations in cortical networks,''
  \emph{science}, vol. 304, no. 5679, pp. 1926--1929, 2004.

\bibitem{wolpert1992stacked}
D.~H. Wolpert, ``Stacked generalization,'' \emph{Neural networks}, vol.~5,
  no.~2, pp. 241--259, 1992.

\bibitem{dlugosz2010realization}
R.~Dlugosz, T.~Talaska, W.~Pedrycz, and R.~Wojtyna, ``Realization of the
  conscience mechanism in {CMOS} implementation of winner-takes-all
  self-organizing neural networks,'' \emph{IEEE Transactions on Neural
  Networks}, vol.~21, no.~6, pp. 961--971, 2010.

\bibitem{BogGriSegHei2016}
B.~Boguslawski, V.~Gripon, F.~Seguin, and F.~Heitzmann, ``Twin neurons for
  efficient real-world data distribution in networks of neural cliques:
  Applications in power management in electronic circuits,'' \emph{IEEE
  transactions on neural networks and learning systems}, vol.~27, no.~2, pp.
  375--387, 2016.

\bibitem{JarOniGriSakSugEndOhnHanGro2014}
H.~Jarollahi, N.~Onizawa, V.~Gripon, N.~Sakimura, T.~Sugibayashi, T.~Endoh,
  H.~Ohno, T.~Hanyu, and W.~J. Gross, ``A non-volatile associative memory-based
  context-driven search engine using 90 nm {CMOS} mtj-hybrid logic-in-memory
  architecture,'' \emph{Journal on Emerging and Selected Topics in Circuits and
  Systems}, vol.~4, pp. 460--474, 2014.

\bibitem{FerGriJia201607}
D.~Ferro, V.~Gripon, and X.~Jiang, ``Nearest neighbour search using binary
  neural networks,'' in \emph{Proceedings of IJCNN}, July 2016.

\bibitem{YuGriJiaJe20153}
C.~Yu, V.~Gripon, X.~Jiang, and H.~J{\'e}gou, ``Neural associative memories as
  accelerators for binary vector search,'' in \emph{COGNITIVE 2015: 7th
  International Conference on Advanced Cognitive Technologies and
  Applications}, 2015, pp. 85--89.

\end{thebibliography}
